\documentclass{article}
\usepackage{iclr2027_conference,times}

\usepackage{amsmath,amsfonts,bm}

\def\eqref#1{Eq.~\ref{#1}}

\def\1{\bm{1}}

\DeclareMathAlphabet{\mathsfit}{\encodingdefault}{\sfdefault}{m}{sl}
\SetMathAlphabet{\mathsfit}{bold}{\encodingdefault}{\sfdefault}{bx}{n}

\usepackage{url}
\usepackage{graphicx}
\usepackage{algorithm}
\usepackage{algpseudocode}
\usepackage{placeins}
\usepackage{booktabs}
\usepackage{array}
\usepackage{xcolor}
\usepackage{hyperref}
\usepackage{xspace}

\newcommand{\ourmethod}{JoHan\xspace}

\newcommand{\bestresult}[1]{\textbf{#1}}
\newcommand{\secondresult}[1]{\underline{#1}}
\newcounter{subtable}[table]

\algrenewcommand{\algorithmicrequire}{\textbf{Input:}}
\algrenewcommand{\algorithmicensure}{\textbf{Output:}}
\newsavebox{\trainingalgorithmbody}
\newsavebox{\inferencealgorithmbody}
\newlength{\pairedalgorithmheight}
\title{Video-Conditioned Generative Joint 2D-3D Hand Motion Recovery}

\author{Chen Xu\textsuperscript{1}, Yunqi Li\textsuperscript{1}, Binbin Huang\textsuperscript{1}, Brent Yi\textsuperscript{2}, Shenghua Gao\textsuperscript{1}, Yi Ma\textsuperscript{1} \\
{\normalfont\textsuperscript{1}The University of Hong Kong}, 
{\normalfont\textsuperscript{2}UC Berkeley}
}

\iclrfinalcopy
\definecolor{preprintcite}{HTML}{447FB3}
\hypersetup{
    colorlinks=true,
    citecolor=preprintcite,
    linkcolor=black,
    urlcolor=black,
    pdftitle={Video-Conditioned Generative Joint 2D-3D Hand Motion Recovery},
    pdfauthor={Chen Xu, Yunqi Li, Binbin Huang, Brent Yi, Shenghua Gao, Yi Ma},
    pdfsubject={Preprint},
    pdfkeywords={hand motion reconstruction, generative modeling, flow matching}
}

\makeatletter
\let\preprint@nmfmt\NAT@nmfmt
\renewcommand{\NAT@nmfmt}[1]{\textcolor{preprintcite}{\preprint@nmfmt{#1}}}
\makeatother

\begin{document}

\maketitle
\lhead{}
\pagestyle{plain}
\thispagestyle{fancy}

\begin{abstract}

Recovering faithful 3D hand motion from video remains challenging due to frequent occlusions and incomplete visual observations, which make frame-wise pose estimates unreliable and temporally inconsistent. To address this problem, we propose \textbf{\ourmethod}, a unified generative framework that recovers hand motion directly from video sequences without relying on intermediate per-frame pose predictions. Trained from scratch, our model jointly generates aligned 2D and 3D \textit{local} hand pose sequences by learning their temporal dynamics and cross-representation correspondence. The generated 2D trajectories exploit direct spatial and temporal cues from the 2D images to guide the following generative 3D motion reconstruction, while the learned motion prior promotes temporal consistency. Their learned 2D--3D correspondence further enables recovery of the hand's {\em global} position and orientation relative to the camera. Extensive experiments on challenging benchmarks demonstrate significantly improved accuracy and speed in local hand-pose and camera-space reconstruction. Notably, our method captures much better hand-motion dynamics, producing significantly smoother motion than previous methods while maintaining high per-frame pose accuracy.

\end{abstract}

\section{Introduction}
\label{sec:introduction}
\vspace{-2mm}
Recovering hand motion from monocular video is a fundamental step toward
understanding how people interact with the physical world. It plays an increasingly
important role in applications such as augmented and virtual reality
(AR/VR), robotics, and human behavior analysis~\citep{human2robot,HRI,rgb2hands,Hoi4D}.
Despite its broad relevance, recovering temporally consistent hand motion remains
challenging. Self-occlusion, occlusion by manipulated objects, 
and motion blur from rapid movement 
reduce the visual evidence available for hand reconstruction.

Many existing approaches formulate hand motion reconstruction as a deterministic
estimation problem. They directly regress hand poses or motion from images or
videos, without explicitly modeling the ambiguities arising from occlusion and
motion blur~\citep{HaMeR,WiLoR,HaWoR,Deformer}.
Generative methods offer a natural way to address these ambiguities by learning a
prior over plausible hand motion and generating pose sequences consistent with
the observations. However, existing generative approaches often still rely on
per-frame pose estimates, either as conditioning signals for motion
generation~\citep{UniHand,HandFlow} or as initial estimates for subsequent
optimization with a learned motion prior~\citep{HMP,Dyn-HaMR}.
These approaches improve temporal consistency, but errors in the observations or initial poses
can still affect the recovered sequence, particularly when occlusion persists.

A related limitation concerns 2D hand motion. 2D hand trajectories provide spatial and temporal evidence for reconstructing 3D motions, and are commonly used as observations~\citep{HMP,HandFlow,WiLoR,UniHand}. However, their own continuity
and dynamics are less explored. Since 2D and 3D motion
are linked through geometric projection, errors and temporal inconsistencies in 2D
trajectories can also compromise 3D motion recovery.

To address these issues, we propose \textbf{\ourmethod}, a unified generative
framework that directly recovers hand motion from video by jointly generating
2D and 3D motion. Generative modeling enables the 3D branch to capture
temporal hand dynamics and account for multiple plausible motions under
ambiguous visual observations. However, 3D motion alone is only weakly
constrained by image evidence due to inherent depth ambiguity. In contrast,
2D trajectories are directly grounded in pixel-space observations, and are capable of providing explicit guidance for
3D reconstruction through their correspondence with 3D joints. Jointly modeling the two modalities therefore combines
the temporal and geometric expressiveness of 3D motion with the strong visual
constraints of 2D trajectories, yielding more accurate 3D hand-motion recovery
than 3D generation alone.

Following HMP~\citep{HMP}, we adopt a local 3D hand-motion representation, decoupling pose dynamics from global orientation and translation. We
similarly normalize 2D motion to remove image-plane location and scale. The
resulting local motions are jointly generated through a unified flow
matching process: at each flow step, our 2D-Guided Fusion uses the
evolving 2D motion to guide 3D generation. Our generators are trained from scratch and conditioned directly on full-frame video features,
while cross-modal supervision promotes geometric consistency between the 2D and 3D
predictions. Finally, we recover global hand motion from the resulting
2D--3D correspondences: after restoring the generated 2D poses
to image coordinates, we use a PnP solver~\citep{EPNP} to estimate the global
orientation and translation of the hand relative to the camera. 

\begin{figure}[t]
    \centering
    \includegraphics[width=\linewidth]{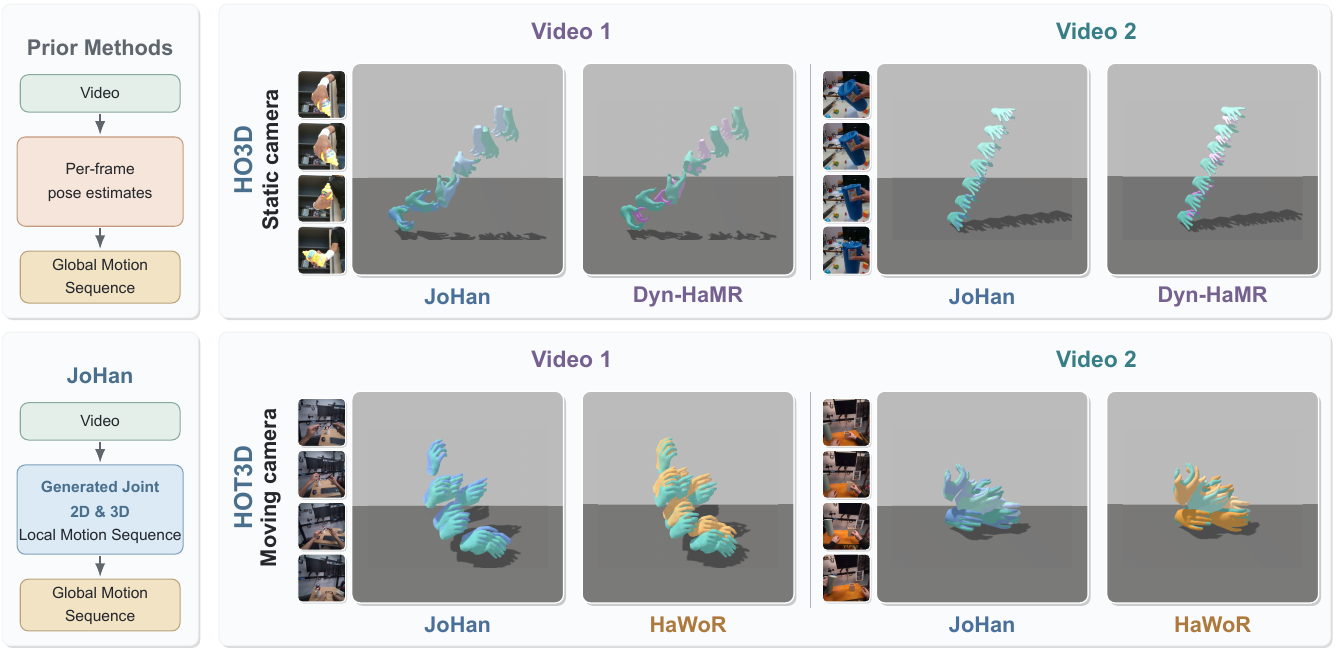}
    \vspace{-4mm}
    \caption{\textbf{Joint motion generation and qualitative comparisons.}
    Left: prior methods typically  use per-frame pose estimates
    to condition or initialize global motion sequence reconstruction. \ourmethod{} jointly
    generates local 2D--3D motion sequences from video, followed by the global reconstruction.
    Right: world-space mesh sequences recovered from two videos from HO3D with a static camera
    (top, compared with \textbf{\textcolor[HTML]{553E6E}{Dyn-HaMR}}~\citep{Dyn-HaMR}) and two from HOT3D with a moving camera
    (bottom, compared with \textbf{\textcolor[HTML]{8C5414}{HaWoR}}~\citep{HaWoR}).
    Methods are denoted by color: \textbf{\textcolor[HTML]{236E69}{Ground truth}}, \textbf{\textcolor[HTML]{2A4A70}{\ourmethod{}}}, \textbf{\textcolor[HTML]{553E6E}{Dyn-HaMR}}, and \textbf{\textcolor[HTML]{8C5414}{HaWoR}}.
    Selected input frames appear in temporal order from top to bottom.\vspace{-4mm}}
    \label{fig:generation_comparison}
\end{figure}
We evaluate \ourmethod{} on HO3Dv2~\citep{HO3D}, HO3Dv3~\citep{HO3D_v3},
and HOT3D~\citep{HOT3D}. On HO3Dv3, it achieves a PA-MPJPE of 7.0\,mm,
a 2D joint error of 4.19 pixels, and the lowest MPJPE among the compared
methods at 32.8\,mm, supporting accurate local reconstruction and geometric
global recovery. On HOT3D, it achieves a PA-MPJPE of 4.75\,mm and an
acceleration error of 2.69\,$\mathrm{m/s^2}$, improving over
HaWoR~\citep{HaWoR} by 21.5\% and 62.3\%, respectively.
As illustrated in Figure~\ref{fig:generation_comparison}, \ourmethod{}
generates temporally coherent and geometrically aligned 2D--3D motion
directly from video, improving local pose and camera-space reconstruction
while producing smoother motion.

Our main contributions are summarized as follows:
\begin{itemize}
    \item We introduce \textbf{\ourmethod}, a unified generative framework that learns a
    hand motion distribution directly from video observations, without relying on
    pretrained per-frame pose estimators.
    \item We jointly generate aligned 2D and 3D local pose sequences within
    a shared motion generation process, modeling their temporal dynamics
    while providing 2D--3D correspondences
    for direct recovery of the hand's global pose in the camera coordinate system.
    \item We demonstrate accurate hand reconstruction on both HO3Dv2 and HO3Dv3, and significantly smoother
    motion with accurate local poses on HOT3D. These results support the
    effectiveness and efficiency of \ourmethod{} across static and moving camera settings.
\end{itemize}

\subsection*{Related Work}
\label{sec:related_work}

\paragraph{Deterministic 3D Hand Estimation.}
Early approaches to 3D hand pose estimation relied on depth observations to
provide geometric cues for model fitting and joint localization
~\citep{DepthKinect,DepthCNN}. The introduction of MANO~\citep{MANO} provided
a compact parametric representation of hand shape and pose, and
\citet{ImageBoukhayma} proposed the first end-to-end framework for recovering
MANO parameters from a single RGB image. More recently, HaMeR~\citep{HaMeR}
improved monocular hand reconstruction by scaling up both transformer-based
architectures and training data, while WiLoR~\citep{WiLoR} integrated hand
localization, mesh reconstruction, and image-aligned refinement into an
end-to-end framework.
Video-based methods further exploit temporal context to improve robustness and
motion consistency. Deformer~\citep{Deformer} fused temporally aligned hand-mesh predictions
to mitigate occlusion and motion blur. HaWoR~\citep{HaWoR} reconstructed temporally coherent hand
motion in camera space and combined it with camera trajectory estimation and
motion infilling to recover complete hand trajectories in world coordinates.
Despite their effectiveness, these methods use regression or model fitting to
produce a single 3D pose or motion trajectory. They therefore cannot explicitly
represent multiple plausible solutions when occlusion, motion blur, or depth
ambiguity leaves the underlying hand motion underconstrained.

2D observations also provide valuable image-space constraints for 3D hand
reconstruction, including reprojection supervision
~\citep{ImageBoukhayma,HaMeR} and geometric refinement. HandTailor
~\citep{HandTailor} refined an initially reconstructed hand mesh through
optimization against image-space evidence, while HandDGP~\citep{HandDGP}
leveraged 2D--3D keypoint correspondences in a differentiable global
positioning module to recover camera-space hand meshes. These methods
demonstrate the value of 2D evidence for spatially aligning 3D hands with
image observations. However, jointly modeling the temporal dynamics of 2D
joint trajectories and their correspondence with 3D hand motion remains
comparatively underexplored.

\paragraph{Generation-Based 3D Hand Reconstruction.}
Generative models represent pose and motion ambiguities through learned
distributions from which plausible hypotheses can be sampled.
Early work focused on single-frame hand poses. Crossing Nets~\citep{CrossingNets} modeled hand poses using a
variational autoencoder (VAE) and depth images using a generative adversarial
network (GAN), connecting the two modalities through a shared latent space. This paradigm was subsequently extended to motion sequences. HMP~\citep{HMP} learned a VAE-based hand-motion prior from
motion-capture data and recovered motion by optimizing its latent representation
to match visual observations. Similarly, Dyn-HaMR~\citep{Dyn-HaMR} incorporated
learned motion priors into a multi-stage optimization framework for reconstructing
interacting hands. These approaches use generative priors to regularize
reconstruction toward plausible and temporally coherent motion, but require
iterative test-time optimization to fit each input sequence.

More recently, conditional generative models have enabled direct hand pose and
motion reconstruction using diffusion or flow matching
~\citep{DDPM,flowmatching}. For individual poses, HHMR~\citep{HHMR} introduced
controllable graph diffusion for hand-mesh generation and reconstruction under
diverse conditions. For motion sequences, UniHand~\citep{UniHand} employed
latent diffusion to generate hand motion conditioned on visual and skeletal
cues, while HandFlow~\citep{HandFlow} applied flow matching to generate
sequences of hand parameters from visual features and 2D skeletal cues. These
skeletal cues provide explicit image-space guidance complementary to visual
features. In their default video-recovery settings, however, both methods obtain their
2D skeletal conditions from frame-wise HaMeR~\citep{HaMeR} predictions, making
motion generation dependent on these estimates while paying limited attention
to the generation and refinement of 2D motion.
In contrast, \ourmethod{} jointly generates local 2D and 3D hand motion through a unified
video-conditioned flow, allowing the evolving 2D trajectories to guide 3D
motion generation without requiring externally estimated poses as explicit
conditioning signals.

\section{Methodology}
\label{sec:methodology}

\subsection{Problem Formulation}
\label{sec:problem_formulation}

\begin{figure}[tbp]
    \centering
    \includegraphics[width=\linewidth]{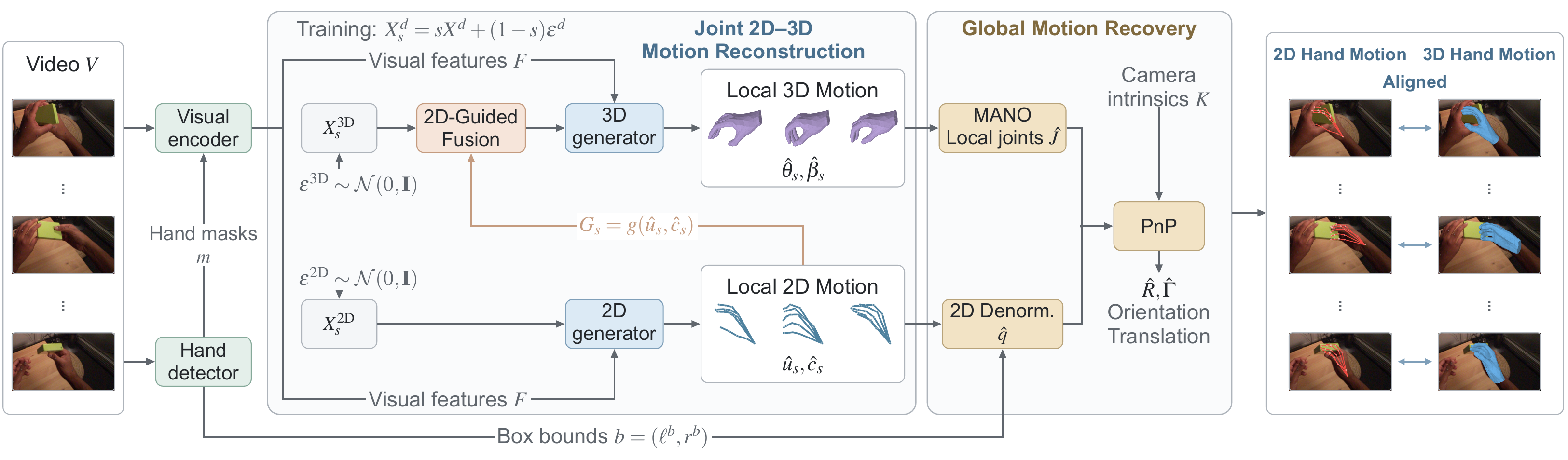}
    \caption{\textbf{Overview of \ourmethod.}
    We first combine full-frame video features with encoded detector-derived hand masks $m$ to form visual conditions $F$. Within the Joint 2D--3D Motion Reconstruction module, the 2D generator predicts clean local motion $\hat{X}^{\mathrm{2D}}_s$ from the noisy state $X^{\mathrm{2D}}_s$ at each flow step. This prediction is then encoded as guidance $G_s$ for the 3D generator, allowing the predicted 2D geometry to inform 3D motion generation. After that, detector-derived bounds restore the image-plane location and scale of the generated 2D local poses. A PnP solver then uses these restored poses, the corresponding local MANO joints, and camera intrinsics $K$ to recover global orientation $\hat{R}$ and translation $\hat{\Gamma}$, which are used to place the reconstructed hands in the camera coordinate system.}
    \label{fig:pipeline}
\end{figure}

Given a monocular video $V=(I_1,\ldots,I_T)$ of $T$ RGB frames,
\ourmethod{} aims to recover the 3D motion of a target hand specified by a
hand-side label $h\in\{\mathrm{left},\mathrm{right}\}$. Each frame
$I_t\in\mathbb{R}^{H\times W\times3}$ has height $H$ and width $W$.
To provide auxiliary cues about the hand's image-plane location and extent,
we represent detector-predicted bounding boxes as binary masks
$m=(m_1,\ldots,m_T)$, where $m_t\in\{0,1\}^{H\times W}$.
As illustrated in Figure~\ref{fig:pipeline}, \ourmethod{}
generates aligned local 2D and 3D hand-motion sequences
$X=(X^{\mathrm{2D}},X^{\mathrm{3D}})$ conditioned on $C=(V,m,h)$,
where $X^d=(x_1^d,\ldots,x_T^d)$ denotes the motion sequence for
$d\in\{\mathrm{2D},\mathrm{3D}\}$, and $x_t^d$ denotes its local state at frame $t$.
The generated 3D local motion is subsequently transformed into the camera coordinate system as a global motion using the global orientation and translation recovered through 2D--3D geometric alignment
(Section~\ref{sec:global_recovery}).

\paragraph{Local Motion Representations.}
We represent 3D hand motion as a sequence of MANO~\citep{MANO} parameters,
comprising hand pose $\theta_t$, shape $\beta_t$, global orientation $R_t$,
and translation $\Gamma_t$ at each frame. Following HMP~\citep{HMP},
we distinguish global and local hand motion. Global orientation and translation
together define the rigid transformation that places the local MANO hand in
the camera coordinate system, whereas removing this transformation leaves
the local motion of the hand joints relative to one another.
The local 3D state is therefore $x_t^{\mathrm{3D}}=(\theta_t,\beta_t)$,
which MANO maps to 3D joints
$J_t=\mathcal{J}(\theta_t,\beta_t)\in\mathbb{R}^{21\times3}$.
Here, $J_{t,j}$ denotes the position of joint $j\in\{1,\ldots,21\}$ at frame
$t$ in the MANO model coordinate system, before applying global orientation
$R_t$ and translation $\Gamma_t$. The corresponding camera-space joints are
denoted by $J_t^{\mathrm{cam}}$.

While global motion is less constrained and varies with the task, body motion,
and camera motion, local motion is governed by hand anatomy and coordinated
joint dynamics. This distinction motivates us to learn a more generalizable
distribution over local motion.

To obtain local 2D poses, we remove the image-plane location and scale of
the hand-joint coordinates $q_{t,j}\in\mathbb{R}^{2}$ in frame $I_t$.
We define their per-axis bounds as $\ell_t^a=\min_j q_{t,j}^a$ and
$r_t^a=\max_j q_{t,j}^a$ for $a\in\{x,y\}$, and compute the normalized
local pose $u_t\in[-1,1]^{21\times2}$ as
\begin{equation}
    u_{t,j}^a=2\frac{q_{t,j}^a-\ell_t^a}{r_t^a-\ell_t^a}-1,
    \qquad u_t=\operatorname{Norm}(q_t;\ell_t,r_t).
    \label{eq:local2d}
\end{equation}
Here, $\operatorname{Norm}$ denotes the per-axis affine normalization above.
The 2D state $x_t^{\mathrm{2D}}=(u_t,v_t)$ also includes joint visibility
labels $v_t\in\{0,1\}^{21}$, for which the model predicts probabilities
$\hat{c}_t\in[0,1]^{21}$.

Using these paired local motion representations,
\ourmethod{} aims to learn a joint prior $p_\phi(X^{\mathrm{2D}},X^{\mathrm{3D}}\mid C)$
under the conditions defined above, with trainable parameters $\phi$.

\subsection{Generation-Based Joint 2D--3D Motion Reconstruction}
\label{sec:joint_generation}

Building on the formulation above, we use flow matching~\citep{flowmatching}
to learn the conditional joint distribution of local 2D and 3D hand motion.
By generating both trajectories from noise under the input conditions, \ourmethod{}
learns their spatial and temporal dependencies directly from video, without
relying on pretrained per-frame pose estimates.

\paragraph{Joint 2D--3D Motion Flow.}
We model local 2D and 3D motion with separately parameterized velocity branches
that evolve a joint motion state along a shared generation timeline.
At generation time $s\in[0,1]$, we define the noisy motion
states by interpolating the paired clean motions with independent Gaussian noises:
\begin{equation}
    X_s^d=sX^d+(1-s)\epsilon^d,
    \qquad \epsilon^d\sim\mathcal{N}(0,\mathbf{I}).
    \label{eq:interpolation}
\end{equation}
Here $d\in\{\mathrm{2D},\mathrm{3D}\}$, and $s$ is shared across both
modalities and all video frames. The paired state
$X_s=(X_s^{\mathrm{2D}},X_s^{\mathrm{3D}})$ connects independent noise at
$s=0$ to aligned clean local motion at $s=1$.

Joint generation alone, however, does not explicitly enforce geometric agreement
between the two predicted motions. While 2D joint trajectories are more directly
observable in the input video, reconstructing 3D motion additionally requires
resolving depth ambiguities. We therefore introduce directional 2D-to-3D guidance
at each flow step. The 2D branch first predicts clean local motion
$\hat{X}_s^{\mathrm{2D}}$, which parameterizes its flow velocity
$\mathbf{v}_{\phi,s}^{\mathrm{2D}}$ as detailed in Eq.~\ref{eq:sampling}.
We encode the same clean prediction as geometric guidance for the 3D branch,
allowing the evolving 2D flow to inform the 3D velocity prediction. Both velocities
then advance their corresponding motion states to the next generation time,
where the guidance is recomputed.

The two branches together define a single joint vector field
$\mathbf{v}_{\phi,s}=(\mathbf{v}_{\phi,s}^{\mathrm{2D}},
\mathbf{v}_{\phi,s}^{\mathrm{3D}})$. Since the 3D branch uses the current
2D prediction, its velocity depends on both motion states. This directional
coupling links the two generation processes at every flow step, allowing the
model to learn a joint motion distribution while encouraging geometric agreement
between the generated 2D and 3D motions.

\paragraph{Joint Flow Parameterization.}
We parameterize the joint flow using two video-conditioned spatiotemporal
generators, $f_{\phi_{\mathrm{2D}}}$ and $f_{\phi_{\mathrm{3D}}}$. Their
clean-motion predictions determine the respective velocity fields
$\mathbf{v}_{\phi,s}^{\mathrm{2D}}$ and
$\mathbf{v}_{\phi,s}^{\mathrm{3D}}$. To condition both generators, we add CNN-encoded hand-mask features to
full-frame patch features from a fine-tuned DINOv3~\citep{DINOv3}, obtaining
$F=(F_t)_{t=1}^{T}$. The masks locate the hand, while full-frame
features retain visual context without fluctuations from detector-defined
crops~\citep{HaMeR,HaWoR,WiLoR}. Following the condition-dropout strategy
of classifier-free guidance~\citep{CFG}, we randomly drop individual frames'
hand-mask conditions during training while retaining video features,
so the generators can use visual evidence when detections are unavailable
or unreliable.

Conditioned on $F$ and hand-side $h$, the 2D generator first predicts clean
local motion from the noisy state
$X_s^{\mathrm{2D}}$ defined in Eq.~\ref{eq:interpolation}.
The encoded 2D guidance $G_s$ then conditions the 3D prediction:
\begin{equation}
    \begin{aligned}
        \hat{X}_s^{\mathrm{2D}}&=f_{\phi_{\mathrm{2D}}}(X_s^{\mathrm{2D}},s,F,h),\\
        \hat{X}_s^{\mathrm{3D}}&=f_{\phi_{\mathrm{3D}}}
        (X_s^{\mathrm{3D}},s,F,h,G_s),
        \qquad G_s=g(\hat{u}_s,\hat{c}_s).
    \end{aligned}
    \label{eq:joint_generator}
\end{equation}
Here the hats denote one-step clean predictions, with
$\hat{X}_s^{\mathrm{2D}}=(\hat{u}_s,\hat{c}_s)$ and
$\hat{X}_s^{\mathrm{3D}}=(\hat{\theta}_s,\hat{\beta}_s)$, while the encoder
$g$ maps the predicted 2D poses and visibility probabilities into guidance
$G_s$ (Eq.~\ref{eq:pose_guidance}). Under the interpolation in
Eq.~\ref{eq:interpolation}, each generator's clean prediction determines
its corresponding velocity component. We advance the paired motion state
through a shared Euler update:
\begin{equation}
    \mathbf{v}_{\phi,s}^d=\frac{\hat{X}_s^d-X_s^d}{1-s},
    \qquad X_{s'}^d=X_s^d+\Delta s\,\mathbf{v}_{\phi,s}^d,
    \label{eq:sampling}
\end{equation}
where $s'=s+\Delta s\leq1$, $s<1$, and $\Delta s$ is the chosen step size.

\paragraph{2D-Guided Fusion.}
We introduce 2D-Guided Fusion to align the generated 2D
guidance with the corresponding 3D motion representation.
At each flow step, we encode the predicted 2D joint coordinates and weight
their features by the predicted visibility probabilities. At frame $t$, we
concatenate the noisy pose parameters of each MANO joint $k$ with guidance
from its corresponding 2D joint:
\begin{equation}
     z_{t,k,s}^{\mathrm{pose}}
    =[\theta_{t,k,s};G_{t,j(k),s}], \qquad G_{t,j,s}=g(\hat{u}_{t,j,s},\hat{c}_{t,j,s})=\hat{c}_{t,j,s}
    E_{\mathrm{2D}}\!\left(\gamma(\hat{u}_{t,j,s})\right),
    \label{eq:pose_guidance}
\end{equation}
Here, $\theta_{t,k,s}$ contains the noisy MANO pose
parameters for joint $k\in\{1,\ldots,15\}$, and $j(k)$ indexes its
corresponding 2D joint. The encoder $g$ applies Fourier encoding $\gamma$
followed by an MLP $E_{\mathrm{2D}}$, with the resulting features weighted
by visibility probability $\hat{c}_{t,j,s}$. The operator
$[\,\cdot\,;\,\cdot\,]$ denotes concatenation, and we project
$z_{t,k,s}^{\mathrm{pose}}$ into a $D$-dimensional motion token. The resulting tokens retain the correspondence between each 2D
guidance feature and its 3D pose joint. We recompute $G_s$ at every step, so the 3D branch uses the evolving 2D geometry throughout generation.

Both generators process $D$-dimensional motion tokens,
with 21 joint tokens per frame in the 2D branch and 15 MANO pose tokens
plus one shape token in the 3D branch. Following the denoiser architecture
of D3DP~\citep{D3DP}, spatial and temporal self-attention model dependencies
across joints and frames, respectively. Each block also uses visual
cross-attention to incorporate $F$, with timestep and hand-side embeddings
supplied through adaptive layer normalization with zero initialization
(AdaLN-Zero)~\citep{DiT}. %

\paragraph{Training Objectives.}
We learn both velocity components by jointly supervising the clean-motion
predictions that parameterize them. During training, we sample
$s=\operatorname{sigmoid}(\xi)$ with $\xi\sim\mathcal{N}(0,1)$ and
evaluate both branches at this generation time. We omit $s$ below for clarity
and use $*$ to denote ground truth, giving the following 2D and 3D losses:
\begin{align}
    \mathcal{L}_{\mathrm{2D}}&=\operatorname{MSE}(\hat{u},u^*)
    +\lambda_c\operatorname{BCE}(\hat{c},v^*),
    \label{eq:loss2d}\\
    \mathcal{L}_{\mathrm{3D}}
    &=\lambda_\theta\operatorname{MSE}(\hat{\theta},\theta^*)
    +\lambda_\beta\operatorname{MSE}(\hat{\beta},\beta^*)
    +\lambda_J\operatorname{MSE}(\hat{J}^{\mathrm{cam}},J^{\mathrm{cam},*}).
    \label{eq:loss3d}
\end{align}
Here, $\hat{J}^{\mathrm{cam}}_t\in\mathbb{R}^{21\times3}$ denotes camera-space
joints obtained from the predicted MANO pose and shape using ground-truth
global orientation $R_t^*$ and translation $\Gamma_t^*$;
$J^{\mathrm{cam},*}_t$ denotes their ground-truth counterparts.
We further couple the two branches through reprojection supervision,
comparing the generated 2D poses with projected 3D joints in the same
normalized coordinate system, with gradients propagating to both branches:
\begin{equation}
    \mathcal{L}_{\mathrm{reproj}}=\operatorname{MSE}\!\left(
    \left\{\operatorname{Norm}\!\left(\Pi_{K_t}
    (\hat{J}^{\mathrm{cam}}_t);\ell_t^*,r_t^*\right)\right\}_{t=1}^{T},\hat{u}\right),
    \label{eq:reprojection}
\end{equation}
where $\Pi_{K_t}$ projects 3D joints to the image plane using camera intrinsics
$K_t$. The bounds $\ell_t^*,r_t^*$ are computed from the ground-truth 2D joints
as in Eq.~\ref{eq:local2d}, so the projected 3D predictions and the 2D targets
$u_t^*$ use the same normalization.

In summary, we train the model by
jointly minimizing
\begin{equation}
    \mathcal{L}=\mathbb{E}
    [\mathcal{L}_{\mathrm{2D}}+\mathcal{L}_{\mathrm{3D}}
    +\lambda_{\mathrm{reproj}}\mathcal{L}_{\mathrm{reproj}}].
    \label{eq:total_loss}
\end{equation}

\subsection{Global Motion Recovery}
\label{sec:global_recovery}

To complete the 3D hand-motion reconstruction, we recover
global orientation and translation from the generated local 2D--3D
correspondences. After restoring the 2D poses to image coordinates, we use
these correspondences and camera intrinsics to solve a perspective-$n$-point
(PnP) problem~\citep{EPNP}.
At inference, the hand detector predicts a bounding box in each frame
for the hand region, and we take its per-axis minimum and maximum coordinates
as the bounds $b_t=(\ell_t^b,r_t^b)$. We correct scale and shift differences
between detected boxes and joint-derived bounds using a lightweight bbox
refiner (Appendix~\ref{app:bbox_refinement}), and apply Kalman
filtering~\citep{Kalman} to suppress frame-to-frame detection jitter.
We use the refined and smoothed bounds to approximate the image-plane
location and scale removed by Eq.~\ref{eq:local2d}, mapping the generated
local poses $\hat{u}_{t,j}$ to image-plane coordinates $\hat{q}_{t,j}$.
We then align the generated local MANO joints $\hat{J}_t$ with the restored
2D coordinates. Let $\hat{o}_t=\hat{J}_{t,\mathrm{wrist}}$ denote the local wrist position
and $\mathcal{I}_t\subseteq\{1,\ldots,21\}$ the joint indices used for
2D--3D alignment. The solver estimates global orientation $\hat{R}_t$
and the camera-space wrist position $\hat{\boldsymbol{\tau}}_t$ by
minimizing the reprojection error
\begin{equation}
    (\hat{R}_t,\hat{\boldsymbol{\tau}}_t)
    =\underset{R\in\mathrm{SO}(3),\,\boldsymbol{\tau}\in\mathbb{R}^3}
    {\operatorname{argmin}}\;
    \sum_{j\in\mathcal{I}_t}
    \left\|\Pi_{K_t}\!\left(R(\hat{J}_{t,j}-\hat{o}_t)+\boldsymbol{\tau}\right)
    -\hat{q}_{t,j}\right\|_2^2.
    \label{eq:pnp}
\end{equation}
Under the MANO convention that applies global orientation about the wrist,
the translation parameter is
$\hat{\Gamma}_t=\hat{\boldsymbol{\tau}}_t-\hat{o}_t$.
Together, the generated local poses and recovered global parameters specify
the full camera-space hand motion.

\begin{table}[t]
    \centering
    \caption{\textbf{Hand reconstruction on HO3D.}
    Published v2 baseline 3D scores; v3 HaMeR/WiLoR re-evaluated by us. 3D errors are in mm;
    $\mathrm{Err}_{\mathrm{2D}}$ is 2D MPJPE in px at $256\times256$.
    Dashes: unavailable. \textbf{Bold}/\underline{underlined}: best/second-best
    before rounding in each HO3D subtable and Table~\ref{tab:hot3d}.}
    \label{tab:ho3d}
    \small
    \setlength{\tabcolsep}{0.5pt}
    \refstepcounter{subtable}\label{tab:ho3d-v2}
    \textbf{(\alph{subtable}) HO3Dv2}\par\smallskip
    \begin{tabular*}{\linewidth}{@{\extracolsep{\fill}}lrrrrrr@{}}
        \toprule
        Method & PA-MPJPE $\downarrow$ & PA-MPVPE $\downarrow$ & AUC$_J$ $\uparrow$ & F@5 $\uparrow$ & F@15 $\uparrow$ & $\mathrm{Err}_{\mathrm{2D}}\downarrow$ \\
        \midrule
        Deformer & 9.4 & 9.1 & -- & 0.546 & 0.963 & -- \\
        HandOccNet & 9.1 & 8.8 & 0.819 & 0.564 & 0.963 & -- \\
        AMVUR & 8.3 & 8.2 & 0.835 & 0.608 & 0.965 & -- \\
        HaMeR & 7.7 & 7.9 & \secondresult{0.846} & 0.635 & \secondresult{0.980} & 6.51 \\
        WiLoR & \secondresult{7.5} & \secondresult{7.7} & \bestresult{0.851} & \secondresult{0.646} & \bestresult{0.983} & \secondresult{6.37} \\
        \midrule
        \ourmethod & \bestresult{7.4} & \bestresult{7.5} & \bestresult{0.851} & \bestresult{0.648} & \bestresult{0.983} & \bestresult{3.92} \\
        \bottomrule
    \end{tabular*}
    \par\medskip
    \refstepcounter{subtable}\label{tab:ho3d-v3}
    \textbf{(\alph{subtable}) HO3Dv3}\par\smallskip
    \begin{tabular*}{\linewidth}{@{\extracolsep{\fill}}lrrrrrrr@{}}
        \toprule
        Method & PA-MPJPE $\downarrow$ & PA-MPVPE $\downarrow$ & MPJPE $\downarrow$ & AUC$_J$ $\uparrow$ & F@5 $\uparrow$ & F@15 $\uparrow$ & $\mathrm{Err}_{\mathrm{2D}}\downarrow$ \\
        \midrule
        HaMeR & 7.4 & 7.3 & 54.7 & 0.851 & 0.643 & 0.978 & \secondresult{6.37} \\
        WiLoR & \secondresult{7.2} & \secondresult{7.1} & \secondresult{53.5} & \secondresult{0.856} & \bestresult{0.654} & \secondresult{0.981} & 6.53 \\
        \midrule
        \ourmethod & \bestresult{7.0} & \bestresult{6.9} & \bestresult{32.8} & \bestresult{0.859} & \secondresult{0.650} & \bestresult{0.984} & \bestresult{4.19} \\
        \bottomrule
    \end{tabular*}
\end{table}

\FloatBarrier
\section{Experiments}
\label{sec:experiments}

\subsection{Experimental Setup}
\label{sec:experimental_setup}
\paragraph{Datasets.}
We train \ourmethod{} on InterHand2.6M~\citep{InterHand26M}, DexYCB~\citep{DexYCB},
ARCTIC~\citep{ARCTIC}, HO3Dv3~\citep{HO3D_v3}, Re:InterHand~\citep{ReInterHand},
and HOT3D~\citep{HOT3D}, excluding held-out evaluation sequences.
We construct clips of up to 128 frames and resize full RGB frames to
$256\times256$; 2D ground-truth joints are obtained by projecting the 3D
annotations with the camera intrinsics.
Appendices~\ref{app:datasets} and~\ref{app:evaluation} detail the data
preprocessing and evaluation protocols, respectively.
We evaluate local pose accuracy on both HO3Dv2 and HO3Dv3 and motion continuity/smoothness
on HOT3D.
For HOT3D evaluation, we use the camera motion estimation module from
HaWoR~\citep{HaWoR} to predict camera trajectories from the input videos.
The main results use DINOv3~\citep{DINOv3} ViT-L/16 as the visual backbone.
We use the YOLO hand detector released with WiLoR~\citep{WiLoR}.
Implementation details are provided in Appendix~\ref{app:implementation}.

\paragraph{Evaluation Metrics.}
We report Procrustes-aligned mean per-joint and mean per-vertex position errors
(PA-MPJPE and PA-MPVPE) to evaluate local hand-pose and mesh accuracy,
respectively. On HO3Dv3, we also report MPJPE, which additionally evaluates camera-space
joint accuracy after global recovery.
Following \citet{HO3D}, we also report the area under the curve
of correctly localized joints (AUC$_J$) and mesh F-scores at 5\,mm and 15\,mm
(F@5, F@15), which assess joint-localization accuracy across thresholds and
mesh reconstruction quality. The 2D joint error $\mathrm{Err}_{\mathrm{2D}}$
measures image-plane pose accuracy in pixels after restoring location and scale.

For world-space motion, we follow \citet{SLAHMR} and use the W-MPJPE and
WA-MPJPE implementations of HaWoR~\citep{HaWoR} to assess global trajectory
accuracy. They align the prediction with ground truth using the first two
frames or the entire evaluated trajectory, respectively. We further report
acceleration error (\mbox{AccEr}) to assess temporal smoothness and agreement
with ground-truth motion dynamics. We provide extensive qualitative results in Appendix~\ref{app:qualitative}.

\subsection{Quantitative Results}
\paragraph{Hand Reconstruction on HO3D.}
HO3Dv3~\citep{HO3D_v3} provides more accurate hand and object pose
annotations than HO3Dv2. Alongside published v2 results, including
HandOccNet~\citep{HandOccNet} and AMVUR~\citep{AMVUR}, we therefore evaluate
the leading v2 baselines HaMeR~\citep{HaMeR} and WiLoR~\citep{WiLoR}
on the official v3 evaluation set.
As shown in Table~\ref{tab:ho3d}, \ourmethod{} achieves the lowest PA-MPJPE and
PA-MPVPE among the compared methods on both versions: 7.4\,mm and 7.5\,mm
on v2, and 7.0\,mm and 6.9\,mm on v3, respectively. These results demonstrate accurate reconstruction of local hand poses and
meshes. For the 2D comparison, we reproject the predicted 3D joints of HaMeR~\citep{HaMeR} and
WiLoR~\citep{WiLoR} into the image. \ourmethod{} also achieves the lowest $\mathrm{Err}_{\mathrm{2D}}$
on both versions, with 3.92 pixels on v2 and 4.19 pixels on v3.
These results support explicitly
learning 2D motion dynamics together with their correspondence to 3D motion.
After recovering global orientation and translation from the generated
correspondences, \ourmethod{} achieves an MPJPE of 32.8\,mm on HO3Dv3,
compared with 53.5\,mm for WiLoR~\citep{WiLoR} and 54.7\,mm for
HaMeR~\citep{HaMeR}. The lower camera-space error, together with the gains
in local pose and 2D joint accuracy, supports the effectiveness of jointly
generating geometrically aligned motions and using their correspondences
for global recovery.

\begin{table}[t]
    \centering
    \begin{minipage}[t]{0.53\linewidth}
    \vspace{0pt}
    \centering
    \caption{\textbf{HOT3D evaluation results.}
    Joint errors are in mm; \mbox{AccEr} is in $\mathrm{m/s^2}$.
    $\dagger$ denotes use of HaWoR's camera motion estimation module;
    Dyn-HaMR uses its own camera pipeline.
    }
    \label{tab:hot3d}
    \vspace{4.28pt}
    {
    \scriptsize
    \selectfont
    \setlength{\tabcolsep}{0.5pt}
    \begin{tabular*}{\linewidth}{@{\extracolsep{\fill}}lrrrr@{}}
        \toprule
        Method & PA-MPJPE $\downarrow$ & W-MPJPE $\downarrow$ & WA-MPJPE $\downarrow$ & AccEr $\downarrow$ \\
        \midrule
        HaMeR$^{\dagger}$ & 9.48 & 135.19 & 33.68 & 12.52 \\
        Dyn-HaMR & 7.48 & 89.19 & 29.13 & 7.41 \\
        WiLoR$^{\dagger}$ & 6.87 & 70.37 & 24.19 & 12.47 \\
        HaWoR & 6.05 & 57.18 & \secondresult{17.76} & 7.14 \\
        HandFlow & 5.49 & \bestresult{43.00} & \bestresult{16.17} & \secondresult{4.18} \\
        UniHand & \secondresult{4.76} & 63.97 & 25.24 & 4.93 \\
        \midrule
        \ourmethod$^{\dagger}$ & \bestresult{4.75} & \secondresult{55.05} & 20.36 & \bestresult{2.69} \\
        \bottomrule
    \end{tabular*}
    \par}
    \end{minipage}\hfill
    \begin{minipage}[t]{0.45\linewidth}
    \vspace{0pt}
    \centering
    \caption{\textbf{Hand recovery efficiency on a 128-frame HOT3D sequence.}
    Parameters count only the main hand-recovery networks.
    Times cover the full hand-recovery pipeline separated as detection and hand pose/motion recovery processes.
     Lower is better for all columns.}
    \label{tab:efficiency}
    {\scriptsize
    \setlength{\tabcolsep}{0.5pt}
    \begin{tabular*}{\linewidth}{@{\extracolsep{\fill}}lrrrr@{}}
        \toprule
        Method & Params. (M) & Detection (s) & Recovery (s) & Total (s) \\
        \midrule 
        WiLoR & \secondresult{630} & \textbf{0.31} & 7.21 & \secondresult{7.52} \\
        HaWoR & 698 & 5.09 & \secondresult{4.26} & 9.35 \\
        HaMeR & 670 & 31.90 & 25.80 & 57.70 \\
        \midrule
        \ourmethod & \textbf{351} & \secondresult{0.34} & \textbf{1.69} & \textbf{2.03} \\
        \bottomrule
    \end{tabular*}
    \par}
    \end{minipage}
\vspace{-3mm}
\end{table}

\paragraph{World-space Motion on HOT3D.}
We evaluate our method on HOT3D-Clips, aligning world-space motion
within each 150-frame clip. We augment HaMeR~\citep{HaMeR} and
WiLoR~\citep{WiLoR} with the camera motion estimation module of
HaWoR~\citep{HaWoR}, which is also used by \ourmethod{} to estimate camera motion.
Each method's own hand predictions are then transformed into world
coordinates using the estimated camera trajectory.
We also include the published results of HandFlow~\citep{HandFlow} and
UniHand~\citep{UniHand} as references. 

As shown in Table~\ref{tab:hot3d},
\ourmethod{} achieves the best local pose accuracy, with a PA-MPJPE of 4.75\,mm.
More notably, its \mbox{AccEr} of 2.69\,$\mathrm{m/s^2}$ is substantially
lower than those of the reproduced baselines, demonstrating smoother
recovered motion while preserving pose accuracy. Together with the HO3D
results, this supports the effectiveness of the learned local motion prior
under different camera-motion conditions, consistent with our motivation
to decouple local hand dynamics from global motion. Nevertheless, our W-MPJPE and WA-MPJPE remain slightly behind the
state-of-the-art results. This limitation reflects our primary focus on local
motion modeling: rather than explicitly learning hand motion in world
coordinates, we recover it by combining the camera-relative hand motion with
the estimated camera motion through a simple geometric transformation.

\paragraph{Computational Efficiency.}
Table~\ref{tab:efficiency} compares parameter counts and inference time on a
single 128-frame HOT3D~\citep{HOT3D} clip, using one NVIDIA H200 GPU. We exclude
camera processing because the compared pipelines share the camera motion
estimation module of HaWoR~\citep{HaWoR}, and separate hand recovery into
hand detection and pose/motion recovery.
\ourmethod{} achieves the lowest total hand recovery time among the compared
methods: 2.03\,s per clip, only 27.0\% of the 7.52\,s required by
WiLoR~\citep{WiLoR}, the fastest baseline. It also uses 351M parameters,
44.3\% fewer than the 630M of this baseline. \ourmethod{} uses a DINOv3~\citep{DINOv3}
ViT-L/16 encoder, which is lighter than the baselines' ViT-H encoders, and
jointly generates the entire motion sequence before geometric global recovery.
These results show that \ourmethod{} combines efficient hand motion recovery
with accurate local poses and smooth motion.

\subsection{Ablations}
\label{sec:ablation}
We conduct all ablations on HO3Dv3 using DINOv3 ViT-B/16 as \ourmethod's visual encoder to reduce
computational cost.
This smaller backbone yields slightly lower accuracy than the main ViT-L/16
model; all variants are compared within the same ablation setting.

\begin{table}[t]
    \centering
    \caption{\textbf{HO3Dv3 ablations.} All variants use DINOv3 ViT-B/16;
    the full model is repeated as the reference for both groups.}
    \label{tab:ablations}
    \fontsize{8}{9}\selectfont
    \setlength{\tabcolsep}{0.5pt}
    \begin{minipage}[t]{0.48\linewidth}
    \vspace{0pt}
    \centering
    \resizebox{0.9\linewidth}{!}{
    \begin{tabular*}{\linewidth}{@{\extracolsep{\fill}}lrr@{}}
        \toprule
        Variant & PA-MPJPE (mm) $\downarrow$ & $\mathrm{Err}_{\mathrm{2D}}$ (px) $\downarrow$ \\
        \midrule
        Full model & 7.8 & 5.21 \\
        \midrule
        \multicolumn{3}{@{}l}{\textit{Auxiliary supervision}} \\
        Without 3D joint loss & 7.9 & 5.32 \\
        Without reprojection loss & 8.2 & 5.96 \\
        \midrule
        \multicolumn{3}{@{}l}{\textit{Use of detections}} \\
        Cropped video & 8.9 & 15.68 \\
        Full video, no masks & 8.0 & $7.8$ \\
        \bottomrule
    \end{tabular*}
    }
    \end{minipage}\hfill
    \begin{minipage}[t]{0.51\linewidth}
    \vspace{0pt}
    \centering
    \resizebox{\linewidth}{!}{
    \begin{tabular*}{\linewidth}{@{\extracolsep{\fill}}lrr@{}}
        \toprule
        Variant & PA-MPJPE (mm) $\downarrow$ & $\mathrm{Err}_{\mathrm{2D}}$ (px) $\downarrow$ \\
        \midrule
        Full model & 7.8 & 5.21 \\
        \midrule
        \multicolumn{3}{@{}l}{\textit{2D--3D generation formulation}} \\
        2D generation only & -- & 5.64 \\
        3D generation only & 8.9 & -- \\
        Separate training & 12.89 & 5.64 \\
        Teacher forcing, noisy 2D & 9.49 & 5.64 \\
        \bottomrule
    \end{tabular*}
    }
    \end{minipage}
\end{table}

\paragraph{Effect of Auxiliary Losses.}
We ablate 3D joint and reprojection supervision in addition to the base
flow matching objective (Table~\ref{tab:ablations}). The 3D joint loss
slightly improves both outputs. Reprojection supervision has a larger
effect, reducing PA-MPJPE from 8.2 to 7.8\,mm and
$\mathrm{Err}_{\mathrm{2D}}$ from 5.96 to 5.21 pixels. These gains support
supervising geometric consistency between generated 2D motion and projected
3D motion to improve reconstruction in both modalities.

\paragraph{Use of Hand Detections.}
Following the crop-based conditioning used by previous
methods~\citep{HaMeR,WiLoR}, we replace full-frame inputs with detected hand
crops. This increases PA-MPJPE from 7.8 to 8.9\,mm and
$\mathrm{Err}_{\mathrm{2D}}$ from 5.21 to 15.68 pixels. Fluctuations in
box size and center may alter apparent hand scale and position after
resizing, disrupting visual continuity, particularly for 2D motion. We also find that 
training and evaluating without hand-mask conditions yields a PA-MPJPE
of 8.0\,mm and an $\mathrm{Err}_{\mathrm{2D}}$ of approximately 7.8 pixels.
Thus, masks provide a modest benefit to local 3D accuracy and a larger
benefit to 2D generation. Accurate local 3D poses can still be recovered
from full frames and the hand-side label alone, supporting our use of
detections as auxiliary conditions for local generation.

\paragraph{Formulation of 2D--3D Generation.}
We first compare joint generation with 2D-only and 3D-only generators.
Compared with the 3D-only generator, \ourmethod{} reduces PA-MPJPE from 8.9 to
7.8\,mm; it also reduces $\mathrm{Err}_{\mathrm{2D}}$ from the 2D-only
generator's 5.64 pixels to 5.21 pixels. These improvements show that jointly
learning the paired motions benefits reconstruction in both modalities.

We further consider training the 2D and 3D generators separately, using
ground-truth 2D motion to condition the 3D generator during training.
The probability chain rule allows the joint distribution to be factorized as
$p(X^{\mathrm{2D}}\mid C)p(X^{\mathrm{3D}}\mid X^{\mathrm{2D}},C)$.
However, this training strategy yields a PA-MPJPE of 12.89\,mm, worse than
the 3D-only generator's 8.9\,mm. This may reflect overreliance on accurate
2D annotations, which are replaced by generated conditions at inference.
Following input-perturbation strategies~\citep{NextX,InputPerturb}, we
perturb ground-truth 2D motion before conditioning the 3D generator in a
noisy teacher-forcing variant. This improves PA-MPJPE to 9.49\,mm but
remains worse than \ourmethod's 7.8\,mm. \ourmethod{} instead uses model-generated
2D guidance during both training and inference, reducing this conditioning
mismatch.

\FloatBarrier
\section{Conclusion}
\label{sec:conclusion}

In this work, we introduced \ourmethod{}, a generative framework for recovering
aligned 2D and 3D hand motion directly from monocular video sequences without relying
on per-frame pose estimates. Our method first jointly generates local 2D and
3D motion, enabling it to capture hand-motion regularities and learn
a generalizable motion prior. Within this joint 2D--3D motion flow, the evolving 2D
prediction guides 3D generation at each step, while reprojection supervision
promotes geometric consistency between the two modalities. The generated local 3D motion
is then transformed into camera coordinates using the global orientation and
translation recovered through 2D--3D geometric alignment.
Experiments on HO3D show that \ourmethod{} accurately reconstructs local hand poses and
meshes and recovers the hand's position and orientation relative to the camera. On HOT3D, it recovers
substantially smoother hand motion while maintaining accurate local poses and high speed.
Our ablation studies further demonstrate that joint generation and
reprojection supervision benefit both 2D and 3D modalities, and that the model retains
accurate local 3D reconstruction even without detection cues. These results
show that jointly learning 2D and 3D motion directly from long sequences improves reconstruction in both modalities and supports temporally coherent and accurate hand motion recovery.

\newpage
\section*{AI Usage Statement}
We used generative AI tools to assist with translation and language
editing, preparation and refinement of figures, and code implementation and execution for selected components of our method. We also used AI tools to help clarify the method's mathematical formulation and present experimental results.

The research ideas and experimental designs were all conceptualized by the
authors, with AI assistance in implementing and running parts of the
experiment. All reported experimental results were obtained
from actual experiments conducted under the authors' supervision.
Generative AI tools were not used to fabricate experimental data or
measurements. The authors have reviewed all AI-assisted content, made necessary
revisions, and take full responsibility for the final content of this work, including its scientific claims, code, figures,
and reported results.


\bibliography{iclr2027_conference}

\begin{thebibliography}{40}
\providecommand{\natexlab}[1]{#1}
\providecommand{\url}[1]{\texttt{#1}}
\expandafter\ifx\csname urlstyle\endcsname\relax
  \providecommand{\doi}[1]{doi: #1}\else
  \providecommand{\doi}{doi: \begingroup \urlstyle{rm}\Url}\fi

\bibitem[Banerjee et~al.(2025)Banerjee, Shkodrani, Moulon, Hampali, Han, Zhang,
  Zhang, Fountain, Miller, Basol, Newcombe, Wang, Engel, and Hodan]{HOT3D}
Prithviraj Banerjee, Sindi Shkodrani, Pierre Moulon, Shreyas Hampali, Shangchen
  Han, Fan Zhang, Linguang Zhang, Jade Fountain, Edward Miller, Selen Basol,
  Richard~A. Newcombe, Robert Wang, Jakob~Julian Engel, and Tomas Hodan.
\newblock {HOT3D:} hand and object tracking in {3D} from egocentric multi-view
  videos.
\newblock In \emph{Proceedings of the IEEE/CVF Conference on Computer Vision
  and Pattern Recognition}, 2025.

\bibitem[Boukhayma et~al.(2019)Boukhayma, de~Bem, and Torr]{ImageBoukhayma}
Adnane Boukhayma, Rodrigo~Andrade de~Bem, and Philip H.~S. Torr.
\newblock {3D} hand shape and pose from images in the wild.
\newblock In \emph{Proceedings of the IEEE/CVF Conference on Computer Vision
  and Pattern Recognition}, 2019.

\bibitem[Chao et~al.(2021)Chao, Yang, Xiang, Molchanov, Handa, Tremblay,
  Narang, Van~Wyk, Iqbal, and Birchfield]{DexYCB}
Yu-Wei Chao, Wei Yang, Yu~Xiang, Pavlo Molchanov, Ankur Handa, Jonathan
  Tremblay, Yashraj~S. Narang, Karl Van~Wyk, Umar Iqbal, and Stan Birchfield.
\newblock {DexYCB}: A benchmark for capturing hand grasping of objects.
\newblock In \emph{Proceedings of the IEEE/CVF Conference on Computer Vision
  and Pattern Recognition}, 2021.

\bibitem[Duran et~al.(2024)Duran, Kocabas, Choutas, Fan, and Black]{HMP}
Enes Duran, Muhammed Kocabas, Vasileios Choutas, Zicong Fan, and Michael~J
  Black.
\newblock {HMP}: Hand motion priors for pose and shape estimation from video.
\newblock In \emph{Proceedings of the IEEE/CVF Winter Conference on
  Applications of Computer Vision}, 2024.

\bibitem[Fan et~al.(2023)Fan, Taheri, Tzionas, Kocabas, Kaufmann, Black, and
  Hilliges]{ARCTIC}
Zicong Fan, Omid Taheri, Dimitrios Tzionas, Muhammed Kocabas, Manuel Kaufmann,
  Michael~J. Black, and Otmar Hilliges.
\newblock {ARCTIC:} {A} dataset for dexterous bimanual hand-object
  manipulation.
\newblock In \emph{Proceedings of the IEEE/CVF Conference on Computer Vision
  and Pattern Recognition}, 2023.

\bibitem[Fu et~al.(2023)Fu, Liu, Xu, Niebles, and Kitani]{Deformer}
Qichen Fu, Xingyu Liu, Ran Xu, Juan~Carlos Niebles, and Kris~M Kitani.
\newblock Deformer: Dynamic fusion transformer for robust hand pose estimation.
\newblock In \emph{Proceedings of the IEEE/CVF International Conference on
  Computer Vision}, 2023.

\bibitem[Ge et~al.(2016)Ge, Liang, Yuan, and Thalmann]{DepthCNN}
Liuhao Ge, Hui Liang, Junsong Yuan, and Daniel Thalmann.
\newblock Robust {3D} hand pose estimation in single depth images: from
  single-view {CNN} to multi-view {CNNs}.
\newblock In \emph{Proceedings of the IEEE/CVF Conference on Computer Vision
  and Pattern Recognition}, 2016.

\bibitem[Hampali et~al.(2020)Hampali, Rad, Oberweger, and Lepetit]{HO3D}
Shreyas Hampali, Mahdi Rad, Markus Oberweger, and Vincent Lepetit.
\newblock {HOnnotate}: A method for {3D} annotation of hand and object poses.
\newblock In \emph{Proceedings of the IEEE/CVF Conference on Computer Vision
  and Pattern Recognition}, 2020.

\bibitem[Hampali et~al.(2021)Hampali, Sarkar, and Lepetit]{HO3D_v3}
Shreyas Hampali, Sayan~Deb Sarkar, and Vincent Lepetit.
\newblock {HO-3D\_v3}: Improving the accuracy of hand-object annotations of the
  {HO-3D} dataset.
\newblock \emph{arXiv preprint arXiv:2107.00887}, 2021.

\bibitem[Ho \& Salimans(2022)Ho and Salimans]{CFG}
Jonathan Ho and Tim Salimans.
\newblock Classifier-free diffusion guidance.
\newblock \emph{arXiv preprint arXiv:2207.12598}, 2022.

\bibitem[Ho et~al.(2020)Ho, Jain, and Abbeel]{DDPM}
Jonathan Ho, Ajay Jain, and Pieter Abbeel.
\newblock Denoising diffusion probabilistic models.
\newblock \emph{Advances in Neural Information Processing Systems}, 2020.

\bibitem[Jiang et~al.(2023)Jiang, Rahmani, Black, and Williams]{AMVUR}
Zheheng Jiang, Hossein Rahmani, Sue Black, and Bryan~M Williams.
\newblock A probabilistic attention model with occlusion-aware texture
  regression for 3d hand reconstruction from a single rgb image.
\newblock In \emph{Proceedings of the IEEE/CVF Conference on Computer Vision
  and Pattern Recognition}, 2023.

\bibitem[Kalman(1960)]{Kalman}
R.~E. Kalman.
\newblock A new approach to linear filtering and prediction problems.
\newblock \emph{Journal of Basic Engineering}, 1960.

\bibitem[Lepetit et~al.(2009)Lepetit, Moreno-Noguer, and Fua]{EPNP}
Vincent Lepetit, Francesc Moreno-Noguer, and Pascal Fua.
\newblock {EPnP}: An accurate {O(n)} solution to the {PnP} problem.
\newblock \emph{Int. J. Comput. Vision}, 2009.

\bibitem[Li et~al.(2024)Li, Zhang, Zhang, Shao, Yu, and Liu]{HHMR}
Mengcheng Li, Hongwen Zhang, Yuxiang Zhang, Ruizhi Shao, Tao Yu, and Yebin Liu.
\newblock {HHMR}: Holistic hand mesh recovery by enhancing the multimodal
  controllability of graph diffusion models.
\newblock In \emph{Proceedings of the IEEE/CVF Conference on Computer Vision
  and Pattern Recognition}, 2024.

\bibitem[Lipman et~al.(2023)Lipman, Chen, Ben{-}Hamu, Nickel, and
  Le]{flowmatching}
Yaron Lipman, Ricky T.~Q. Chen, Heli Ben{-}Hamu, Maximilian Nickel, and Matthew
  Le.
\newblock Flow matching for generative modeling.
\newblock In \emph{International Conference on Learning Representations}, 2023.

\bibitem[Liu et~al.(2022)Liu, Liu, Jiang, Lyu, Wan, Shen, Liang, Fu, Wang, and
  Yi]{Hoi4D}
Yunze Liu, Yun Liu, Che Jiang, Kangbo Lyu, Weikang Wan, Hao Shen, Boqiang
  Liang, Zhoujie Fu, He~Wang, and Li~Yi.
\newblock {HOI4D:} {A} {4D} egocentric dataset for category-level human-object
  interaction.
\newblock In \emph{Proceedings of the IEEE/CVF Conference on Computer Vision
  and Pattern Recognition}, 2022.

\bibitem[Lv et~al.(2021)Lv, Xu, Yang, Qian, Mao, and Lu]{HandTailor}
Jun Lv, Wenqiang Xu, Lixin Yang, Sucheng Qian, Chongzhao Mao, and Cewu Lu.
\newblock {HandTailor}: Towards high-precision monocular {3D} hand recovery.
\newblock In \emph{Proceedings of the British Machine Vision Conference}, 2021.

\bibitem[Moon et~al.(2020)Moon, Yu, Wen, Shiratori, and Lee]{InterHand26M}
Gyeongsik Moon, Shoou{-}I Yu, He~Wen, Takaaki Shiratori, and Kyoung~Mu Lee.
\newblock {InterHand2.6M}: {A} dataset and baseline for {3D} interacting hand
  pose estimation from a single {RGB} image.
\newblock In Andrea Vedaldi, Horst Bischof, Thomas Brox, and Jan{-}Michael
  Frahm (eds.), \emph{European Conference on Computer Vision}, 2020.

\bibitem[Moon et~al.(2023)Moon, Saito, Xu, Joshi, Buffalini, Bellan, Rosen,
  Richardson, Mize, de~Bree, Simon, Peng, Garg, McPhail, and
  Shiratori]{ReInterHand}
Gyeongsik Moon, Shunsuke Saito, Weipeng Xu, Rohan Joshi, Julia Buffalini,
  Harley Bellan, Nicholas Rosen, Jesse Richardson, Mallorie Mize, Philippe
  de~Bree, Tomas Simon, Bo~Peng, Shubham Garg, Kevyn McPhail, and Takaaki
  Shiratori.
\newblock A dataset of relighted {3D} interacting hands.
\newblock In \emph{Advances in Neural Information Processing Systems}, 2023.

\bibitem[Ning et~al.(2023)Ning, Sangineto, Porrello, Calderara, and
  Cucchiara]{InputPerturb}
Mang Ning, Enver Sangineto, Angelo Porrello, Simone Calderara, and Rita
  Cucchiara.
\newblock Input perturbation reduces exposure bias in diffusion models.
\newblock In Andreas Krause, Emma Brunskill, Kyunghyun Cho, Barbara Engelhardt,
  Sivan Sabato, and Jonathan Scarlett (eds.), \emph{International Conference on
  Machine Learning}, 2023.

\bibitem[Oikonomidis et~al.(2011)Oikonomidis, Kyriazis, and
  Argyros]{DepthKinect}
Iason Oikonomidis, Nikolaos Kyriazis, and Antonis Argyros.
\newblock Efficient model-based {3D} tracking of hand articulations using
  {Kinect}.
\newblock In \emph{Proceedings of the British Machine Vision Conference}, 2011.

\bibitem[Park et~al.(2022)Park, Oh, Moon, Choi, and Lee]{HandOccNet}
JoonKyu Park, Yeonguk Oh, Gyeongsik Moon, Hongsuk Choi, and Kyoung~Mu Lee.
\newblock {HandOccNet}: Occlusion-robust {3D} hand mesh estimation network.
\newblock In \emph{Proceedings of the IEEE/CVF Conference on Computer Vision
  and Pattern Recognition}, 2022.

\bibitem[Pavlakos et~al.(2024)Pavlakos, Shan, Radosavovic, Kanazawa, Fouhey,
  and Malik]{HaMeR}
Georgios Pavlakos, Dandan Shan, Ilija Radosavovic, Angjoo Kanazawa, David
  Fouhey, and Jitendra Malik.
\newblock Reconstructing hands in {3D} with transformers.
\newblock In \emph{Proceedings of the IEEE/CVF Conference on Computer Vision
  and Pattern Recognition}, 2024.

\bibitem[Peebles \& Xie(2023)Peebles and Xie]{DiT}
William Peebles and Saining Xie.
\newblock Scalable diffusion models with transformers.
\newblock In \emph{Proceedings of the IEEE/CVF International Conference on
  Computer Vision}, 2023.

\bibitem[Potamias et~al.(2025)Potamias, Zhang, Deng, and Zafeiriou]{WiLoR}
Rolandos~Alexandros Potamias, Jinglei Zhang, Jiankang Deng, and Stefanos
  Zafeiriou.
\newblock {WiLoR}: End-to-end {3D} hand localization and reconstruction
  in-the-wild.
\newblock In \emph{Proceedings of the IEEE/CVF Conference on Computer Vision
  and Pattern Recognition}, 2025.

\bibitem[Qi et~al.(2024)Qi, Ma, Cui, and Yu]{HRI}
Jing Qi, Li~Ma, Zhenchao Cui, and Yushu Yu.
\newblock Computer vision-based hand gesture recognition for human-robot
  interaction: a review.
\newblock \emph{Complex \& Intelligent Systems}, 2024.

\bibitem[Ren et~al.(2025)Ren, Yu, He, Shen, Yuille, and Chen]{NextX}
Sucheng Ren, Qihang Yu, Ju~He, Xiaohui Shen, Alan~L. Yuille, and Liang{-}Chieh
  Chen.
\newblock Beyond next-token: {Next-X} prediction for autoregressive visual
  generation.
\newblock In \emph{Proceedings of the IEEE/CVF International Conference on
  Computer Vision}, 2025.

\bibitem[Romero et~al.(2017)Romero, Tzionas, and Black]{MANO}
Javier Romero, Dimitrios Tzionas, and Michael~J. Black.
\newblock Embodied hands: Modeling and capturing hands and bodies together.
\newblock \emph{ACM Transactions on Graphics (TOG)}, 2017.

\bibitem[Shan et~al.(2023)Shan, Liu, Zhang, Wang, Han, Wang, Ma, and Gao]{D3DP}
Wenkang Shan, Zhenhua Liu, Xinfeng Zhang, Zhao Wang, Kai Han, Shanshe Wang,
  Siwei Ma, and Wen Gao.
\newblock Diffusion-based {3D} human pose estimation with multi-hypothesis
  aggregation.
\newblock In \emph{Proceedings of the IEEE/CVF International Conference on
  Computer Vision}, 2023.

\bibitem[Sim{\'{e}}oni et~al.(2026)Sim{\'{e}}oni, Vo, Seitzer, Baldassarre,
  Oquab, Jose, Khalidov, Szafraniec, Yi, Ramamonjisoa, Massa, Haziza,
  Wehrstedt, Wang, Darcet, Moutakanni, Sentana, Roberts, Vedaldi, Tolan,
  Brandt, Couprie, Mairal, J{\'{e}}gou, Labatut, and Bojanowski]{DINOv3}
Oriane Sim{\'{e}}oni, Huy~V. Vo, Maximilian Seitzer, Federico Baldassarre,
  Maxime Oquab, Cijo Jose, Vasil Khalidov, Marc Szafraniec, Seung~Eun Yi,
  Micha{\"{e}}l Ramamonjisoa, Francisco Massa, Daniel Haziza, Luca Wehrstedt,
  Jianyuan Wang, Timoth{\'{e}}e Darcet, Th{\'{e}}o Moutakanni, Leonel Sentana,
  Claire Roberts, Andrea Vedaldi, Jamie Tolan, John Brandt, Camille Couprie,
  Julien Mairal, Herv{\'{e}} J{\'{e}}gou, Patrick Labatut, and Piotr
  Bojanowski.
\newblock {DINOv3}.
\newblock \emph{Trans. Mach. Learn. Res.}, 2026.

\bibitem[Sun et~al.(2026)Sun, Wu, Tu, Dong, and Wu]{UniHand}
Zhihao Sun, Tong Wu, Ruirui Tu, Daoguo Dong, and Zuxuan Wu.
\newblock {UniHand}: A unified model for diverse controlled {4D} hand motion
  modeling.
\newblock In \emph{International Conference on Learning Representations}, 2026.

\bibitem[Valassakis \& Garcia-Hernando(2024)Valassakis and
  Garcia-Hernando]{HandDGP}
Eugene Valassakis and Guillermo Garcia-Hernando.
\newblock {HandDGP}: Camera-space hand mesh prediction with differentiable
  global positioning.
\newblock In \emph{European Conference on Computer Vision}, 2024.

\bibitem[Wan et~al.(2017)Wan, Probst, Van~Gool, and Yao]{CrossingNets}
Chengde Wan, Thomas Probst, Luc Van~Gool, and Angela Yao.
\newblock {Crossing Nets}: Combining {GANs} and {VAEs} with a shared latent
  space for hand pose estimation.
\newblock In \emph{Proceedings of the IEEE/CVF Conference on Computer Vision
  and Pattern Recognition}, 2017.

\bibitem[Wang et~al.(2020)Wang, Mueller, Bernard, Sorli, Sotnychenko, Qian,
  Otaduy, Casas, and Theobalt]{rgb2hands}
Jiayi Wang, Franziska Mueller, Florian Bernard, Suzanne Sorli, Oleksandr
  Sotnychenko, Neng Qian, Miguel~A. Otaduy, Dan Casas, and Christian Theobalt.
\newblock {RGB2Hands}: real-time tracking of {3D} hand interactions from
  monocular {RGB} video.
\newblock \emph{ACM Transactions on Graphics (TOG)}, 2020.

\bibitem[Xie et~al.(2025)Xie, Cao, Weng, Xing, Chen, Shen, Leng, Wu, and
  Jiang]{human2robot}
Sicheng Xie, Haidong Cao, Zejia Weng, Zhen Xing, Haoran Chen, Shiwei Shen,
  Jiaqi Leng, Zuxuan Wu, and Yu-Gang Jiang.
\newblock {Human2Robot}: Learning robot actions from paired human-robot videos.
\newblock In \emph{Proceedings of the AAAI Conference on Artificial
  Intelligence}, 2025.

\bibitem[Xu et~al.(2026)Xu, Duan, Gu, Shen, Xu, and Yue]{HandFlow}
Mingxi Xu, Bowen Duan, Yi~Gu, Zhengyang Shen, Renjing Xu, and Yutao Yue.
\newblock {HandFlow}: Fully generative {4D} hand recovery with flow matching.
\newblock \emph{arXiv preprint arXiv:2607.11221}, 2026.

\bibitem[Ye et~al.(2023)Ye, Pavlakos, Malik, and Kanazawa]{SLAHMR}
Vickie Ye, Georgios Pavlakos, Jitendra Malik, and Angjoo Kanazawa.
\newblock Decoupling human and camera motion from videos in the wild.
\newblock In \emph{Proceedings of the IEEE/CVF Conference on Computer Vision
  and Pattern Recognition}, 2023.

\bibitem[Yu et~al.(2025)Yu, Zafeiriou, and Birdal]{Dyn-HaMR}
Zhengdi Yu, Stefanos Zafeiriou, and Tolga Birdal.
\newblock {Dyn-HaMR}: Recovering {4D} interacting hand motion from a dynamic
  camera.
\newblock In \emph{Proceedings of the IEEE/CVF Conference on Computer Vision
  and Pattern Recognition}, 2025.

\bibitem[Zhang et~al.(2025)Zhang, Deng, Ma, and Potamias]{HaWoR}
Jinglei Zhang, Jiankang Deng, Chao Ma, and Rolandos~Alexandros Potamias.
\newblock {HaWoR}: World-space hand motion reconstruction from egocentric
  videos.
\newblock In \emph{Proceedings of the IEEE/CVF Conference on Computer Vision
  and Pattern Recognition}, 2025.

\end{thebibliography}
\bibliographystyle{iclr2027_conference}

\clearpage
\appendix
\section{Appendix}
\label{app:appendix}

\begin{figure}[H]
    \centering
    \includegraphics[width=\linewidth]{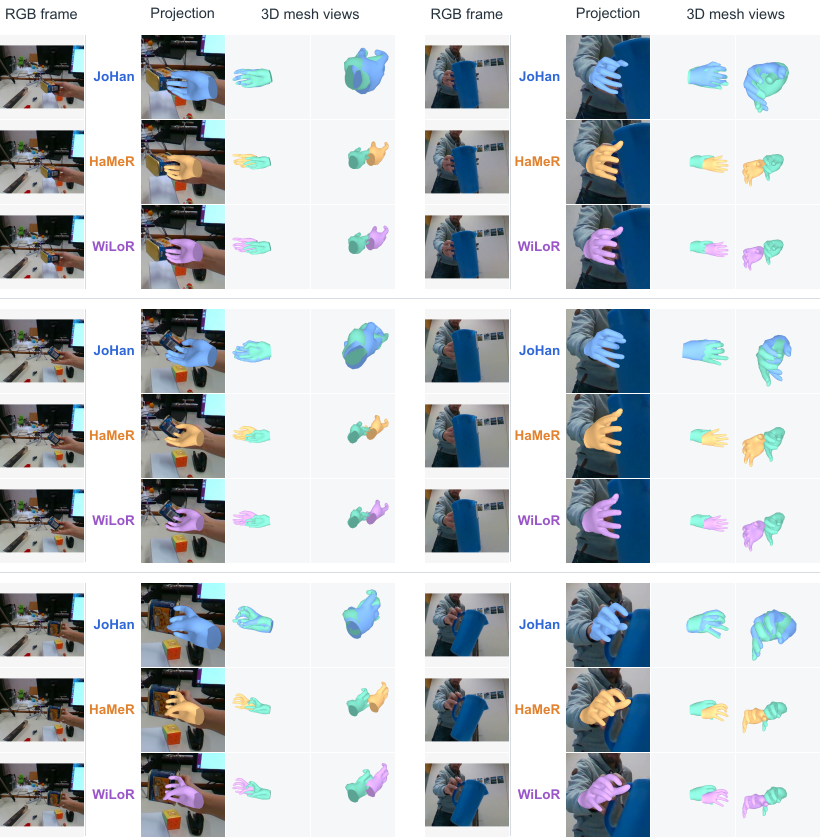}
    \caption{\textbf{Qualitative hand reconstruction from videos in HO3D.}
    Each group shows one selected frame. Each method's row contains the
    input RGB frame, its predicted mesh projected onto the hand crop,
    and two camera-space mesh views overlaid with ground truth.
    Ground truth is green; \ourmethod{}, HaMeR~\citep{HaMeR}, and
    WiLoR~\citep{WiLoR} are blue, orange, and purple, respectively.}
    \label{fig:qualitative_ho3d}
\end{figure}

\clearpage

\begin{figure}[H]
    \centering
    \includegraphics[width=\linewidth]{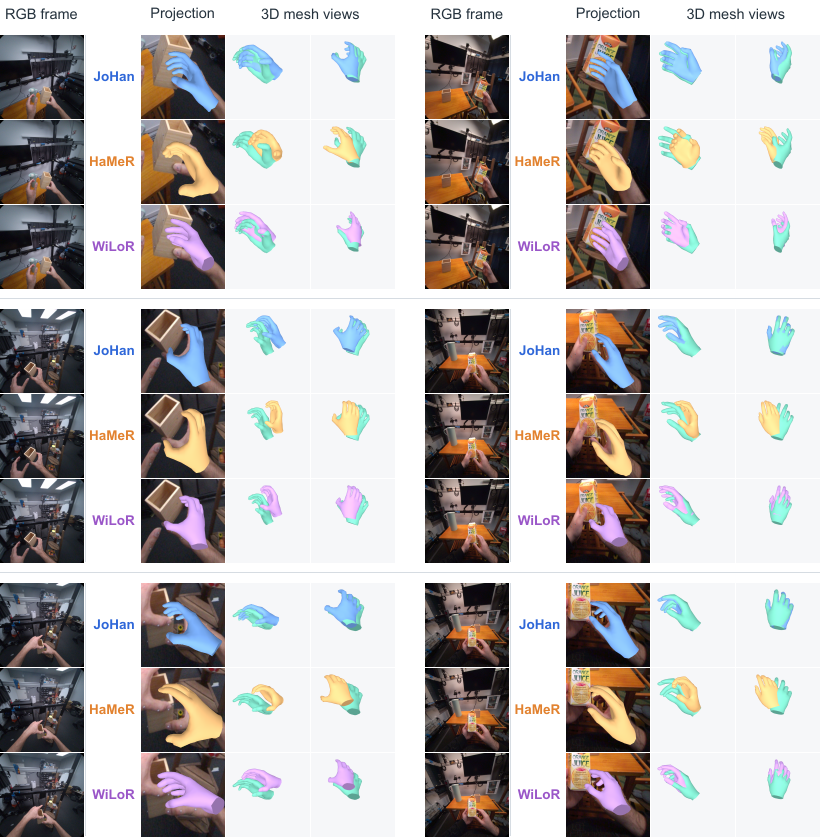}
    \caption{\textbf{Qualitative hand reconstruction from videos in HOT3D.}
    We show six selected frames with hand--object occlusion.
    Each row shows the input RGB frame, a mesh projection onto the hand
    crop, and two camera-space mesh views overlaid with ground truth.
    The three method rows compare \ourmethod{},
    HaMeR~\citep{HaMeR}, and WiLoR~\citep{WiLoR}.
    The layout and colors follow Figure~\ref{fig:qualitative_ho3d}.}
    \label{fig:qualitative_hot3d}
\end{figure}

\clearpage

\begin{figure}[H]
    \centering
    \includegraphics[width=0.94\linewidth]{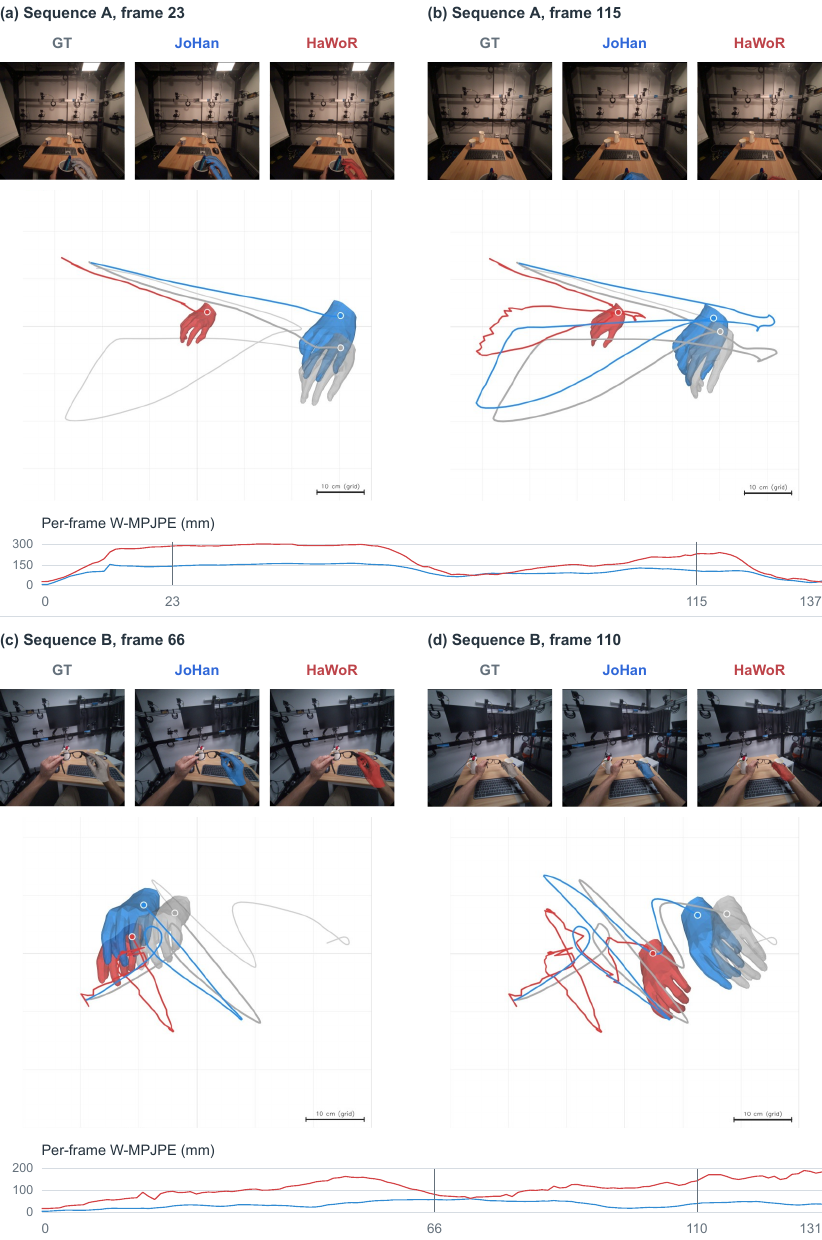}
    \caption{\textbf{World-space motion comparison on HOT3D.}
    Four examples from two sequences compare ground truth (gray),
    \ourmethod{} (blue), and HaWoR~\citep{HaWoR} (red).
    The enlarged views project initially aligned world-space meshes and
    trajectories onto the first two ground-truth principal axes.
    Grid spacing is 10\,cm. Each row shows two frames from one sequence,
    indexed from zero. The curves show per-frame W-MPJPE; vertical
    markers indicate the displayed frames.}
    \label{fig:qualitative_trajectories}
\end{figure}

\clearpage

\begin{figure}[H]
    \centering
    \includegraphics[width=\linewidth]{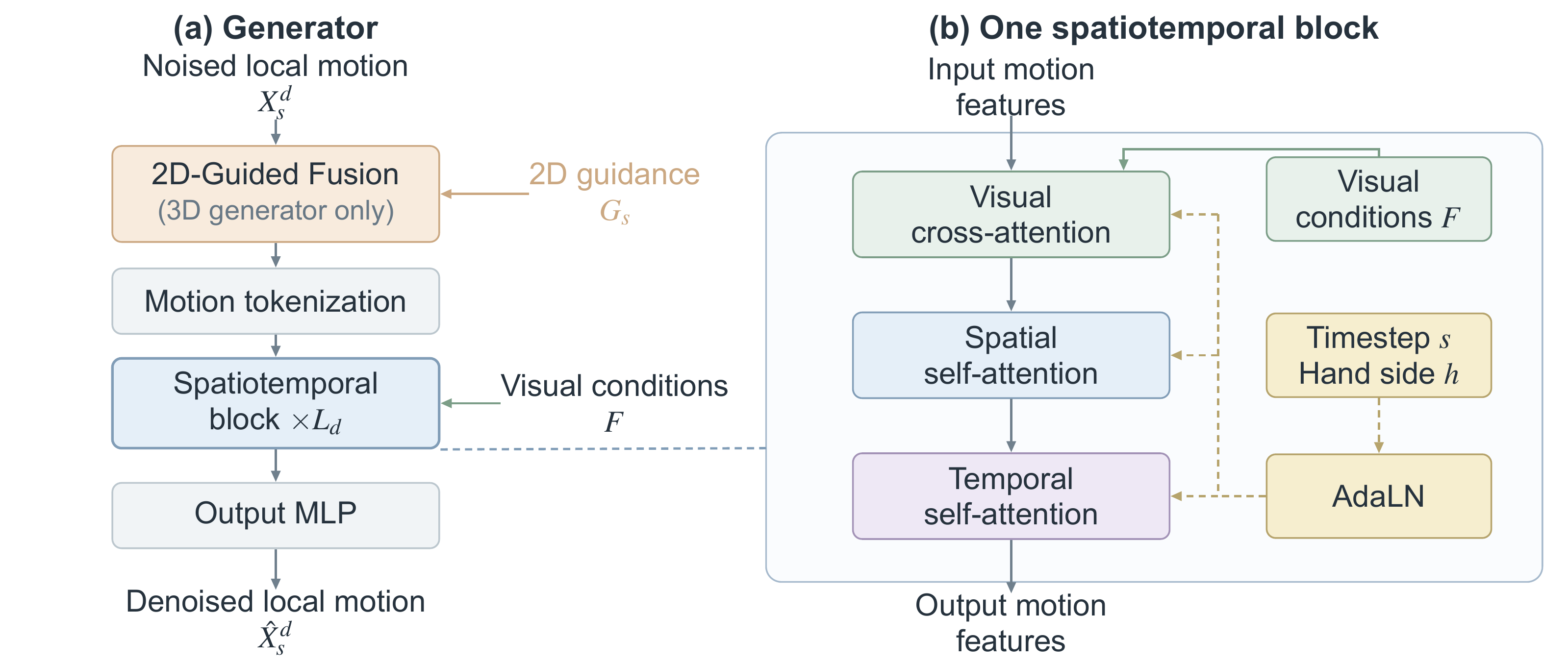}
    \caption{\textbf{Conditioned generator architecture.}
    (a) Each generator maps noised local motion $X_s^d$ to a clean-motion
    prediction $\hat{X}_s^d$. Only the 3D generator uses 2D-Guided Fusion,
    which incorporates the guidance $G_s$ from the current 2D prediction
    before motion tokenization.
    (b) Each spatiotemporal block applies visual cross-attention conditioned
    on $F$, followed by spatial and temporal self-attention. Gold dashed
    arrows indicate AdaLN-Zero conditioning from timestep $s$ and hand
    side $h$ in both generators.}
    \label{fig:generator}
\end{figure}

\subsection{Implementation Details}
\label{app:implementation}

\paragraph{Training and inference.}
We train \ourmethod{} on 32 GPUs for 300,000 iterations with an effective
batch size of 256 clips. Each GPU processes two clips per forward pass,
and gradients are accumulated over four passes before each optimizer update.
The two motion generators are trained from scratch, while the pretrained
DINOv3~\citep{DINOv3} ViT-L/16 backbone is fine-tuned jointly with them.
We use AdamW with learning rates $\eta_{\mathrm{gen}}=2\times10^{-4}$ for the generators
and $\eta_{\mathrm{vis}}=2\times10^{-5}$ for the visual backbone,
and 1,000 warm-up iterations. We maintain an exponential moving average
of the model parameters with decay $\rho=0.9999$ for inference.
For the training objective in Eq.~\ref{eq:total_loss}, we set
$\lambda_c=\lambda_\theta=\lambda_{\mathrm{reproj}}=1$,
$\lambda_\beta=0.1$, and $\lambda_J=100$.
Following the condition-dropout strategy of classifier-free
guidance~\citep{CFG}, hand-mask conditions are dropped independently per
frame with probability 0.3 during training; the RGB video remains available.
At inference, both motion states start from independent Gaussian noise
and are updated through $10$ joint flow steps, with the 2D prediction
recomputed before the 3D prediction at every step.

\paragraph{Generator architecture.}
Figure~\ref{fig:generator} illustrates the architecture shared by the two
generators, and Table~\ref{tab:app_architecture} lists their dimensions.
The 2D generator embeds the noisy joint coordinates and visibility states
into 21 motion tokens per frame. The 3D generator uses 15 MANO pose tokens
and one token for the ten shape parameters. Both branches use a token
dimension of $D=384$, but have independent parameters and different depths:
four spatiotemporal blocks for 2D and six for 3D.
Each block first applies visual cross-attention, followed by spatial
self-attention within a frame and temporal self-attention across frames.
Attention layers are accompanied by feedforward layers and residual
connections. We use GELU activations and rotary positional embeddings
in the spatial and temporal self-attention layers. Timestep $s$ and
hand-side $h$ embeddings condition both generators through
AdaLN-Zero~\citep{DiT}, which modulates normalized features and gates
residual updates. Separate output heads predict 2D coordinates and
visibility probabilities in the 2D branch, and MANO pose and shape
parameters in the 3D branch.

\begin{table}[htbp]
    \centering
    \caption{\textbf{Motion generator configurations.}
    Both generators use the same full-frame visual conditions and
    timestep/hand-side conditioning, with independent attention parameters.}
    \label{tab:app_architecture}
    \begin{tabular}{@{}lcc@{}}
        \toprule
        Component & 2D generator & 3D generator \\
        \midrule
        Motion tokens per frame & 21 & 15 pose + 1 shape \\
        Token dimension & 384 & 384 \\
        Spatiotemporal blocks & 4 & 6 \\
        Attention heads & 4 & 4 \\
        Feedforward hidden dimension & 384 & 1536 \\
        2D-Guided Fusion & -- & $G_s$ at each flow step \\
        Output per frame & 21 2D joints + visibility & MANO pose + shape \\
        \bottomrule
    \end{tabular}
\end{table}

\paragraph{Visual conditioning and 2D guidance.}
For each $256\times256$ RGB frame, DINOv3 ViT-L/16 produces a $16\times16$
grid of 1024-dimensional patch features. Binary hand masks are pooled to
this grid and then projected to 1024 dimensions. We add the resulting mask features to the
image features and project them to the generators' token dimension.
The spatial patch grid is retained for visual cross-attention.
In the 3D branch, 2D-Guided Fusion encodes the current predicted 2D
coordinates with Fourier features and an MLP, then weights them by
predicted visibility. As in Eq.~\ref{eq:pose_guidance}, we concatenate each
guidance feature with the noisy pose parameters of its corresponding MANO
joint, then project the concatenated vector into a motion token.
Guidance from the wrist and fingertips also contributes to the shape
token. This guidance is generated by the 2D branch during both training
and inference.

\paragraph{Training and inference algorithms.}
Algorithms~\ref{alg:training} and~\ref{alg:inference} summarize the two
procedures. Here, $\mathcal{E}_\phi$ denotes the visual encoder and mask
feature fusion, and $\phi$ includes all trainable encoding and generation
parameters. Training clips provide paired local targets
$X^*=(X^{\mathrm{2D},*},X^{\mathrm{3D},*})$ and joint-derived bounds
$b^*=(\ell^*,r^*)$ through the preprocessing in Appendix~\ref{app:datasets}.
Training times and noises are sampled independently for each clip,
with the same $s$ used for its two modalities.
The modality index $d\in\{\mathrm{2D},\mathrm{3D}\}$ applies to both
branches, and $\hat{X}^{\mathrm{2D}}=(\hat{u},\hat{c})$ and
$\hat{X}^{\mathrm{3D}}=(\hat{\theta},\hat{\beta})$ denote their predictions.
At inference, $\phi$ denotes the trained EMA weights, and the schedule
$\mathcal{S}=\{(s_i,s_{i+1}-s_i)\}_{i=0}^{N-1}$ uses
$s_i=\sin(\frac{\pi i}{2N})$ with $N=10$.
Box refinement, denormalization, MANO evaluation, and PnP are applied
framewise, while Kalman filtering operates on the box trajectory.

\begin{figure}[htbp]
\begin{lrbox}{\trainingalgorithmbody}
\begin{minipage}[t]{0.49\linewidth}
    \footnotesize
    \begin{algorithmic}[1]
        \Require Training set $\mathcal{D}$ of videos $V$, masks $m$, hand
        sides $h$, local targets $X^*$, joint-derived bounds $b^*$,
        camera-space joints $J^{\mathrm{cam},*}$, global parameters
        $(R^*,\Gamma^*)$, and intrinsics $K$; initial parameters $\phi$;
        number of updates $M$; generator and visual-backbone learning
        rates $\eta_{\mathrm{gen}},\eta_{\mathrm{vis}}$; EMA decay $\rho$
        \Ensure Model parameters $\phi$ and EMA weights $\bar\phi$
        \State $\bar\phi\gets\phi$
        \For{$i=1,\ldots,M$}
            \State Sample a minibatch $\mathcal{B}$ from $\mathcal{D}$
            \State Drop each frame's mask in $m$ with probability 0.3 to obtain $\tilde m$
            \State $F\gets\mathcal{E}_\phi(V,\tilde m)$
            \State $\xi\sim\mathcal{N}(0,1)$; $s\gets\operatorname{sigmoid}(\xi)$
            \State $\epsilon^d\sim\mathcal{N}(0,\mathbf I)$ independently
            \State $X_s^d\gets sX^{d,*}+(1-s)\epsilon^d$
            \State $\hat X_s^{\mathrm{2D}}\gets f_{\phi_{\mathrm{2D}}}(X_s^{\mathrm{2D}},s,F,h)$
            \State $G_s\gets g(\hat u_s,\hat c_s)$
            \State $\hat X_s^{\mathrm{3D}}\gets f_{\phi_{\mathrm{3D}}}(X_s^{\mathrm{3D}},s,F,h,G_s)$
            \State Compute minibatch loss $\widehat{\mathcal L}$ from
            Eq.~\ref{eq:total_loss}, using $b^*$ for reprojection normalization
            \State Update $\phi$ with AdamW using $\nabla_\phi\widehat{\mathcal L}$
            and learning rates $\eta_{\mathrm{gen}},\eta_{\mathrm{vis}}$
            \State $\bar\phi\gets\rho\bar\phi+(1-\rho)\phi$
        \EndFor
        \State \Return $(\phi,\bar\phi)$
    \end{algorithmic}
\end{minipage}
\end{lrbox}
\begin{lrbox}{\inferencealgorithmbody}
\begin{minipage}[t]{0.49\linewidth}
    \footnotesize
    \begin{algorithmic}[1]
        \Require Video $V$, hand-side $h$, intrinsics $K$; trained EMA
        parameters $\phi$, hand detector, and bbox refiner; flow schedule
        $\mathcal{S}$
        \Ensure Local motions $\hat X^{\mathrm{2D}},\hat X^{\mathrm{3D}}$;
        image-space joints $\hat q$; global orientation $\hat R$ and
        MANO translation $\hat\Gamma$
        \State Detect target-hand boxes $b$ from $(V,h)$
        \State Rasterize $b$ into binary hand masks $m$
        \State $F\gets\mathcal{E}_\phi(V,m)$
        \State $X_0^d\sim\mathcal{N}(0,\mathbf I)$ independently
        \For{$(s,\Delta s)\in\mathcal S$ in order}
            \State $\hat X_s^{\mathrm{2D}}\gets f_{\phi_{\mathrm{2D}}}(X_s^{\mathrm{2D}},s,F,h)$
            \State $G_s\gets g(\hat u_s,\hat c_s)$
            \State $\hat X_s^{\mathrm{3D}}\gets f_{\phi_{\mathrm{3D}}}(X_s^{\mathrm{3D}},s,F,h,G_s)$
            \State $X_{s+\Delta s}^d\gets X_s^d+\frac{\Delta s}{1-s}(\hat X_s^d-X_s^d)$
        \EndFor
        \State $\hat X^d\gets X_1^d$
        \State Refine $b$ using $(V,b)$, then apply Kalman filtering
        \State $(\ell^b,r^b)\gets b$
        \State $\hat q\gets\ell^b+\frac{\hat u+1}{2}(r^b-\ell^b)$
        \State $\hat J\gets\mathcal J(\hat\theta,\hat\beta)$; $\hat o\gets\hat J_{\mathrm{wrist}}$
        \State $(\hat R,\hat{\boldsymbol\tau})\gets
        \operatorname{PnP}(\hat J-\hat o,\hat q,K)$ \Comment{Eq.~\ref{eq:pnp}}
        \State $\hat\Gamma\gets\hat{\boldsymbol\tau}-\hat o$
        \State \Return $(\hat X^{\mathrm{2D}},\hat X^{\mathrm{3D}},\hat q,\hat R,\hat\Gamma)$
    \end{algorithmic}
\end{minipage}
\end{lrbox}
\setlength{\pairedalgorithmheight}{\dimexpr\ht\trainingalgorithmbody+\dp\trainingalgorithmbody\relax}
\ifdim\dimexpr\ht\inferencealgorithmbody+\dp\inferencealgorithmbody\relax>\pairedalgorithmheight
    \setlength{\pairedalgorithmheight}{\dimexpr\ht\inferencealgorithmbody+\dp\inferencealgorithmbody\relax}
\fi
\begin{minipage}[t]{0.49\linewidth}
\vspace{0pt}
\begin{algorithm}[H]
    \caption{\strut Training \ourmethod}
    \label{alg:training}
    \vbox to \pairedalgorithmheight{\hbox{\usebox{\trainingalgorithmbody}}\vfil}
\end{algorithm}
\end{minipage}\hfill
\begin{minipage}[t]{0.49\linewidth}
\vspace{0pt}
\begin{algorithm}[H]
    \caption{\strut Inference with \ourmethod}
    \label{alg:inference}
    \vbox to \pairedalgorithmheight{\hbox{\usebox{\inferencealgorithmbody}}\vfil}
\end{algorithm}
\end{minipage}
\end{figure}

\FloatBarrier
\subsection{Qualitative Results}
\label{app:qualitative}

\paragraph{Pose and mesh reconstruction.}
Figures~\ref{fig:qualitative_ho3d} and~\ref{fig:qualitative_hot3d}
compare \ourmethod{} with HaMeR~\citep{HaMeR} and WiLoR~\citep{WiLoR}
on six selected frames from each dataset. The image overlays show
hand placement relative to the observed hand and manipulated object,
while the additional mesh views reveal differences in finger poses
and 3D alignment with ground truth. These examples illustrate that
\ourmethod{} recovers detailed hand poses under hand--object occlusion
while preserving their placement in the image.

\paragraph{World-space motion on HOT3D.}
Figure~\ref{fig:qualitative_trajectories} compares \ourmethod{} with
HaWoR~\citep{HaWoR} using four examples from two HOT3D sequences.
The two panels in each row show different time points from the same
sequence, with enlarged world-space views beneath the image overlays.
In these examples, \ourmethod{} follows smoother paths with fewer
abrupt changes than HaWoR, consistent with the lower acceleration
errors reported in Tables~\ref{tab:hot3d} and~\ref{tab:hot3d_gt_boxes}.
The accompanying per-frame W-MPJPE curves provide a separate view
of world-space reconstruction accuracy over each sequence.

\subsection{Data Preprocessing}
\label{app:datasets}

\paragraph{Data sources and common representation.}
We combine the training data from InterHand2.6M~\citep{InterHand26M},
DexYCB~\citep{DexYCB}, ARCTIC~\citep{ARCTIC}, HO3D~\citep{HO3D_v3},
Re:InterHand~\citep{ReInterHand}, and HOT3D~\citep{HOT3D}.
The egocentric and external camera streams of ARCTIC are processed
separately, and left- and right-hand tracks are treated as independent
motion sequences. We convert annotations into a common camera-space
representation containing MANO parameters, 21 hand joints, hand-side label,
camera intrinsics, and temporal validity masks. Geometric quantities
are expressed in meters. Held-out evaluation sequences are excluded
from training before temporal clips are constructed.

\paragraph{Image and annotation preprocessing.}
We resize full RGB frames to $256\times256$ without hand-centered
cropping and update camera intrinsics consistently. Egocentric ARCTIC
and Re:InterHand images are undistorted using the provided calibration.
For the Aria RGB stream of HOT3D, we undistort the images and apply the
image-orientation convention used by HaWoR~\citep{HaWoR}, including a
$90^\circ$ clockwise rotation, before resizing.

We use a common SMPL-X MANO implementation across datasets. Where
source shape conventions differ, we convert the shape parameters to
this implementation and adjust translation to preserve the wrist
position. Camera-space joint annotations are retained for image
projection, while the ground-truth joints used for MANO-based 3D
supervision are computed from the annotated parameters with the same
hand model used for predictions. All 2D joint targets are obtained by
projecting the stored 3D annotations using the camera intrinsics.
The per-axis minima and maxima of these projected joints define the
ground-truth bounds $\ell_t^*,r_t^*$ for local 2D normalization and
reprojection supervision. These training bounds are distinct from the
detector-derived bounds used for denormalization at inference.

\paragraph{Training clip construction.}
We split single-hand tracks at sequence boundaries and annotation
discontinuities. Tracks with fewer than 32 valid frames are discarded.
Tracks of 32--128 frames form one clip; longer tracks are divided into
128-frame windows with a stride of 16 frames, with an additional
endpoint-aligned window when needed to cover the end of the track.
We trim leading and trailing frames without finite projected joints
inside the image and discard clips that become shorter than 32 frames.
The retained clips are padded to 128 frames, and temporal validity
masks exclude padded frames from supervision. This construction
preserves temporal ordering while supporting variable-length
observations within a common training format.

\subsection{Evaluation Details}
\label{app:evaluation}

\paragraph{HO3D.}
We evaluate on the official evaluation sets of HO3Dv2~\citep{HO3D}
and HO3Dv3~\citep{HO3D_v3}, using the released ground-truth 3D joints
and mesh vertices for each version. We use the official evaluation
code to compute PA-MPJPE, PA-MPVPE, AUC$_J$, F@5, and F@15.
These reference joints and vertices are used directly, rather than
being regenerated from MANO parameters with our training implementation.
For the additional MPJPE evaluation on HO3Dv3, we compare the recovered
camera-space joints with the released joint annotations without
Procrustes alignment, thereby assessing reconstruction after global
orientation and translation recovery. We compute $\mathrm{Err}_{\mathrm{2D}}$
against the projected ground-truth joints in the $256\times256$ image
coordinate system, with camera intrinsics adjusted for resizing.

\paragraph{HOT3D.}
We use HOT3D-Clips~\citep{HOT3D}, which provides selected subsequences
from the original recordings, each containing approximately 150 frames
or five seconds of video. We use the Aria subset and condition on its
main-view RGB images; the Quest3 recordings do not provide RGB images.
The annotated clips include ground-truth hand poses and camera
trajectories for world-space evaluation.

The official test split does not release ground-truth annotations.
As in HaWoR~\citep{HaWoR}, UniHand~\citep{UniHand}, and
HandFlow~\citep{HandFlow}, we therefore evaluate on held-out official
training data. Specifically, we adopt the 27 Aria recordings held out
by HaWoR~\citep{HaWoR}, performing the split at the source-video level before clip
construction. These recordings are excluded from all model training.
Our training pipeline constructs windows from the remaining full
recordings, as described in Appendix~\ref{app:datasets}.

Our processed HOT3D-Clips index contains 1,167 training clips and 349
test clips, with no overlap in their source videos. The frame mapping
used for the main evaluation selects 333 test clips from 25 of the
27 held-out recordings. After annotation-validity and shared detection-mask
filtering, 42,990 right-hand frames contribute to the \ourmethod{} results
in Table~\ref{tab:hot3d}. These evaluation clips are distinct from the
128-frame windows used for training. For \ourmethod{}, the camera motion
estimation module of HaWoR~\citep{HaWoR} supplies only the estimated
camera trajectories used to transform the reconstructed camera-space
hand motion into world coordinates. The additional ground-truth-box
experiment uses 100-frame evaluation segments, as detailed in
Appendix~\ref{app:hot3d_gt_boxes}.

\subsection{Bounding-Box Refinement for Global Recovery}
\label{app:bbox_refinement}

\paragraph{Correcting the denormalization bounds.}
Detector boxes identify the hand region but need not tightly enclose
the projected hand joints. Their centers and extents can therefore
differ from the joint-derived bounds used to normalize training
poses. Directly using these boxes for denormalization introduces
scale and shift errors in the recovered image-space 2D joints, which
can also affect the global orientation and translation estimated
by PnP. To reduce this mismatch, we train a lightweight bounding-box
refiner after the detector to predict residual scale and center
corrections.

During training, we perturb ground-truth joint-derived boxes in
scale and position and supervise the refiner to recover their original
bounds. The refiner combines CNN features from an image region around
the perturbed box with an MLP embedding of its box coordinates, and
predicts a log-scale correction and a two-dimensional center offset.
The scale correction rescales the box extent, while the center offset
is expressed relative to the input box size. At inference, the same
network corrects detector-predicted boxes before they are used to
restore the image-plane location and scale of the generated 2D motion.
The refiner is trained separately from the motion generators and
requires only a few hours of training. The generators continue to
receive full-frame video features.

\paragraph{Effect of bounding-box refinement.}
Table~\ref{tab:bbox_refiner} compares reconstruction with and without
the refiner on the same HOT3D evaluation frames. Refinement reduces
image-space 2D error from 5.18 to 3.27\,px and camera-space MPJPE from
46.05 to 33.84\,mm. It also improves world-space trajectory accuracy,
reducing W-MPJPE from 78.37 to 55.05\,mm. These results support
correcting the image-plane location and scale used for denormalization
before geometric global recovery. The refiner predicts box corrections
rather than directly updating the generated MANO pose parameters.
To further examine motion smoothness without the influence of
detector-box correction, we additionally evaluate with ground-truth
boxes below.

\begin{table}[htbp]
    \centering
    \caption{\textbf{Bounding-box refiner ablation on HOT3D.}
    Both settings are evaluated on the same 42,990 frames.
    Joint errors are in mm, AccEr is in $\mathrm{m/s^2}$, and
    $\mathrm{Err}_{\mathrm{2D}}$ is in pixels.
    Lower values are better for all metrics.}
    \label{tab:bbox_refiner}
    \small
    \setlength{\tabcolsep}{3pt}
    \begin{tabular*}{\linewidth}{@{\extracolsep{\fill}}lrrrrrr@{}}
        \toprule
        Bbox refiner & PA-MPJPE $\downarrow$ & MPJPE $\downarrow$ & W-MPJPE $\downarrow$ & WA-MPJPE $\downarrow$ & AccEr $\downarrow$ & $\mathrm{Err}_{\mathrm{2D}}$ $\downarrow$ \\
        \midrule
        Without & 5.20 & 46.05 & 78.37 & 26.86 & 3.42 & 5.18 \\
        With & 4.75 & 33.84 & 55.05 & 20.36 & 2.69 & 3.27 \\
        \bottomrule
    \end{tabular*}
\end{table}

\subsection{HOT3D Evaluation with Ground-Truth Bounding Boxes}
\label{app:hot3d_gt_boxes}

The main HOT3D evaluation uses detected hand regions. To examine
whether our smoothness advantage persists without detector-box
preprocessing, we also evaluate \ourmethod{} and the reproduced
baselines using ground-truth hand boxes. Following the evaluation
setting of HaWoR~\citep{HaWoR}, we use 100-frame segments for this
experiment. Ground-truth boxes remove detector localization errors
and the need to refine or smooth predicted boxes.
Table~\ref{tab:hot3d_gt_boxes} reports the resulting
local pose, world-space trajectory, and acceleration errors.

\begin{table}[htbp]
    \centering
    \caption{\textbf{HOT3D results with ground-truth bounding boxes.}
    All rows report our evaluations on 100-frame segments using ground-truth hand boxes.
    Joint errors are in mm and AccEr is in $\mathrm{m/s^2}$.
    \ourmethod{} uses HaWoR's camera motion estimation module.
    Lower values are better for all metrics.}
    \label{tab:hot3d_gt_boxes}
    \begin{tabular*}{\linewidth}{@{\extracolsep{\fill}}lrrrr@{}}
        \toprule
        Method & PA-MPJPE $\downarrow$ & W-MPJPE $\downarrow$ & WA-MPJPE $\downarrow$ & AccEr $\downarrow$ \\
        \midrule
        HaMeR & 9.00 & 145.15 & 36.31 & 11.07 \\
        WiLoR & 6.28 & 72.81 & 24.72 & 9.48 \\
        HaWoR & 5.47 & 42.95 & 13.43 & 5.13 \\
        \midrule
        \ourmethod & 4.70 & 45.78 & 14.89 & 3.34 \\
        \bottomrule
    \end{tabular*}
\end{table}

The conclusions are consistent with the detection-based evaluation:
\ourmethod{} combines accurate local poses with lower acceleration
error. Compared with HaWoR~\citep{HaWoR}, \ourmethod{} achieves a lower
PA-MPJPE of 4.70\,mm versus 5.47\,mm. Its W-MPJPE and WA-MPJPE are
comparable, though slightly higher: 45.78\,mm versus 42.95\,mm and
14.89\,mm versus 13.43\,mm, respectively. At the same time, its AccEr
decreases from 5.13 to 3.34\,$\mathrm{m/s^2}$, and is also lower than
those of HaMeR~\citep{HaMeR} and WiLoR~\citep{WiLoR}. Since this
experiment uses shorter evaluation segments than the main HOT3D-Clips
comparison, we assess the methods within Table~\ref{tab:hot3d_gt_boxes}.
The lower acceleration error under this setting shows that
\ourmethod{} retains its smoothness advantage without detector-box
preprocessing, providing additional evidence for temporally coherent
reconstruction with the learned motion prior.

\end{document}